\documentclass[acmsmall,nonacm]{acmart}
\usepackage{booktabs}
\usepackage{multirow}
\usepackage{xcolor}
\usepackage{graphicx}
\usepackage{float}
\newcommand{\rev}[1]{\textcolor{red}{#1}}
\newcommand{\revb}[1]{\textcolor{red}{#1}}
\renewcommand{\rev}[1]{#1}
\renewcommand{\revb}[1]{#1}
\newcommand{\revblue}[1]{\textcolor{blue}{#1}} \renewcommand{\revblue}[1]{#1}% 2026-11-04: marking off for submission
\hypersetup{pdftitle={Same Error, Different Recovery: How Delivery Changes LLM Agent Behavior},colorlinks=false}

\begin{document}
\raggedbottom

\title{\rev{Same Error, Different Recovery:\\How Delivery \revb{Changes} LLM Agent Behavior}}

\author{Sunghyeun Woo}
\orcid{0009-0000-2709-1262}
\email{nunconnect1@gmail.com}
\affiliation{%
  \institution{Independent Researcher}
  \city{Bucheon}
  \country{Republic of Korea}}
  
\author{Javier Conde}
\orcid{0000-0002-5304-0626}
\email{javier.conde.diaz@upm.es}
\affiliation{\department{Information Processing and Telecommunications Center}\institution{Universidad Polit\'{e}cnica de Madrid}\city{Madrid}\country{Spain}}

\author{Pedro Reviriego}
\authornote{Corresponding author.}
\orcid{0000-0003-2540-5234}
\email{pedro.reviriego@upm.es}
\affiliation{%
  \department{Information Processing and Telecommunications Center}
  \institution{Universidad Polit\'{e}cnica de Madrid}
  \city{Madrid}
  \country{Spain}}

\begin{abstract}
\rev{A tool-using language model must decide whether to retry a failed call, correct it, or stop and tell the user. The error text alone may not be enough: the same feedback can lead to different actions depending on where it appears in the conversation. We isolate this delivery problem by varying placement, guidance, retained history and turn structure. Two studies with plans fixed before data collection, across 13 open models from 3B to 70B parameters, compare the next action and subsequent task completion, with exact deterministic reruns. A new user turn that retains the failed exchange avoids the severe stopping failure observed in Llama models when that exchange is removed. It is non-inferior to history rewriting in pooled synthetic comparisons, but is not best for every model. We implement this form as an opt-in \revb{recovery layer within a deterministic inference engine} and evaluate it on public benchmarks and a coding agent. On ToolMaze, it raises success on explicit errors by 14.2 points with 7--8B models and 10.5 points with 32--70B models; \revb{Loud/Quiet, 24 multi-turn tasks with visible failures and silent corruption, scored by final-state agreement,} shows no measurable gain. A client given the exact fault kind performs better still. The findings support delivery as a distinct factor in recovery, not an engine-exclusive solution or a universal retry policy.}
\end{abstract}


\maketitle
\hypersetup{pdftitle={Same Error, Different Recovery: How Delivery Changes LLM Agent Behavior}}

%% ===================================================================
\section{Introduction}
\rev{A failed tool call creates a practical decision for an LLM agent: try again, change the call, or stop. Consider three illustrative cases drawn from the error types used in this paper. A data request returns HTTP~503: a retry may work. An edit fails because \texttt{old\_string} is empty: the arguments must change. A data service returns HTTP~403 because a token is required: repeating the request cannot supply the missing permission. These cases require different next actions, even though each reaches the model as a tool result.}

\rev{The problem is not only what the error says, but how the model receives it. An error can remain buried in a tool message, be restated as a runtime note, or appear in a new user turn. Removing the failed exchange may discard useful evidence; retaining it may change whether the model stops. We ask whether these delivery choices change recovery when the feedback content is held fixed. In our measurements, after a transient error left in the tool message, five of six open 7--35B models repeated the call in at most 37\% of cases, and one of them in none.}

\rev{Prior work shows that feedback content and the way the failed call is recorded can change recovery \citep{gumaan2026feedback,ray2026structured,panthi2026outcome}, and that the role in which identical content arrives changes whether a model flags a reasoning error \citep{chen2026self}. We study delivery for tool failures: where the feedback appears, whether the failed exchange remains, and whether a new turn follows. Guidance explicitly states a desired next action, so correct-action scores measure response to that guidance, not unaided diagnosis. Comparisons using identical guidance estimate the additional effect of a delivery form, which can change more than one of these elements at once (Section~\ref{sec:dimensions}).}

\rev{We implement the intervention in an inference engine because it already renders the conversation for the model. \revb{A sufficiently capable client or gateway could perform the same message transformation.} The engine implementation provides one common, opt-in path across clients, but must infer failures from flags and text.}

\rev{The paper makes the following contributions.}
\begin{enumerate}
\item \rev{A controlled design space for error delivery that varies placement, guidance, retained history and turn structure.}
\item \rev{Two studies across 13 open models from 3B to 70B parameters, with plans fixed before data collection (Study~2's small tier in a public registry) and exact reruns. They show that delivery affects the next action and that keeping the failed exchange while adding a new user turn avoids a severe stopping failure in Llama models. Benefits vary across models and settings.}
\item \rev{An opt-in implementation in \revb{a deterministic inference engine} and an evaluation against llama.cpp, vLLM, SGLang and two client-side alternatives. The layer improves ToolMaze recovery on explicit errors, has no measurable effect on Loud/Quiet, and improves reporting of access failures in a small coding-agent evaluation.}
\end{enumerate}

\rev{The argument proceeds from the delivery problem and implementation (Section~\ref{sec:methods}), to experimental protocols (Section~\ref{sec:experiments}), to results (Section~\ref{sec:results}). The three studies answer successive questions: which delivery components matter, whether retaining history avoids the observed failure, and whether the implemented layer helps in tool-using tasks. The main text explains what each comparison establishes; the appendices retain the complete numerical comparisons, negative results and deviations.}

The engine, including the tool-error layer, is open source \citep{woo2026geulpu}, and the records of all three studies (plans and registrations, the hash lists that fixed them, stimuli, scripts, raw outputs and analyses) are archived on Zenodo \citep{woo2026study}. Appendices~A--I are provided as supplementary material.

%% ===================================================================
\section{Background and Related Work}
\label{sec:related}

\paragraph{Tool-using agents.}
LLM agents interleave reasoning with actions in an environment \citep{yao2023react} and learn or are prompted to call external tools \citep{schick2023toolformer}.
Function calling is now evaluated on multi-step, stateful tasks \citep{patil2025the} and on interactions among tools, agents and users \citep{yao2024taubench}.
Coding agents such as Qwen Code \citep{qwenlm2025qwen} run long loops of file and shell tools against open-weight models \citep{qwen2025qwen25,yang2025qwen3,grattafiori2024the}, usually served by an inference engine such as vLLM \citep{kwon2023efficient}, SGLang \citep{zheng2024sglang} or llama.cpp \citep{ggmlorg2023llamacpp}.

\paragraph{Tool failures in benchmarks.}
A growing set of benchmarks injects tool failures on purpose.
Loud/Quiet \citep{kraishan2026loud} adds timeouts, missing tools, schema drift and silent corruption to BFCL multi-turn tasks and finds three or more consecutive calls to the same tool in up to 22.2\% of trials.
ToolMaze \citep{zhu2026when} crosses explicit and implicit failures with transient and permanent ones.
ParaRecover \citep{guan2026pararecover} evaluates error localization and recovery in parallel tool use, and CRITICTOOL \citep{huang2025critictool} evaluates whether models can critique their own tool-calling errors.
Live environments produce failures without injection: MCP-Atlas \citep{bandi2026mcpatlas} runs tasks against production MCP servers that return real rate limits and schema mismatches, and ARE \citep{froger2025are} gives each tool a random failure probability in Gaia2.
AgentChaos \citep{tan2026agentchaos} injects faults one layer lower, at the HTTP interface to the LLM API, and reports pass@1 drops of up to 50 points.
An interaction-centric taxonomy places each agent failure mode on an edge between two components, such as the model and its harness \citep{raj2026model}.
Bench2Robust \citep{chen2026bench2robust} adds tool-failure episodes to existing benchmarks that call for retrying, switching to an equivalent tool or recognizing that no path remains, and supplies structured recovery context at run time. These benchmarks measure robustness, and some add recovery context; none of them compares deliveries of the same failure text.

\paragraph{\revb{How the failure is described.}}
Several studies vary the failure text or how the failed call is recorded.
In small models (135M--1.7B parameters), a record of the failed call raises the probability of repeating it from 0.06 to 0.54, a runtime-generated description of the failure removes 76\% of that effect, and retrying from a clean context that drops the failed attempt repeats the call most often \citep{gumaan2026feedback}.
In an agent loop whose candidate actions are checked by external validators, structured feedback that names admissible alternatives more than doubles repair rates, from 14 to 36 of 50 tasks for one model \citep{ray2026structured}.
Outcome monitors attach an in-band receipt to the returned tool result that names the violated property and the available recovery tools; this raises ToolMaze success from 10.9\% to 28.1\% across four models \citep{panthi2026outcome}.
In text-to-Cypher agents, raw database errors work about as well as schema-grounded feedback, and detecting the failure and retrying is what matters \citep{prokopiou2026what}.
Typed tool interfaces reduce interface misuse in a pilot study \citep{sigdel2026schema}, and an audit of MCP servers finds that typed fields flag 18 of 21 induced failures but never state a cause or a repair \citep{mehan2026can}.
An engineering report on terminal coding agents describes injecting recovery guidance per error type and observes, in early experiments, that reminders sent in the system role work worse than reminders in the user role \citep{bui2026building}.
Retrying is not free: blind retries of tools with side effects duplicate writes \citep{li2026where}, a caveat we return to in Section~\ref{sec:discussion}.
\rev{Closest to our question, role relabeling keeps an erroneous claim byte-identical and moves it among the thought, user, tool and system-memory roles; moving it out of the thought role raises explicit correction by 23 to 93 points in math and logic tasks \citep{chen2026self}. It concerns flagging a reasoning error, whereas we study the next action after a failed tool call (retry, fix or stop), whether the failed exchange is kept, and later task completion. Apart from the recording of the failed call \citep{gumaan2026feedback} and its role \citep{chen2026self}, these studies primarily examine feedback content within a particular harness. ReIn is a related inference-time intervention: an external module diagnoses conversational errors and inserts recovery plans into the agent's reasoning \citep{kim2026rein}. Its intervention changes the recovery content and reasoning context. Our controlled form comparisons instead keep the guidance fixed, and among C2--C4 also the note, while changing placement, retained history and turn structure.}

\paragraph{\revb{Recovery location.}}
Recovery is currently the client's job.
Tool protocols let the client mark a failure: Anthropic's \texttt{is\_error} \citep{anthropic2026handle} and MCP's \texttt{isError}, which clients should pass to the model \citep{modelcontextprotocol2026tools}.
Agent frameworks add handlers on top: Google ADK's Reflect-and-Retry plugin retries a failed tool up to three times with reflection guidance \citep{google2026reflect}, LangChain provides tool-error and tool-retry middleware \citep{langchain2026prebuilt}, the OpenAI Agents SDK turns a tool exception into a message to the model through a configurable function \citep{openai2026tools}, and Claude Code lets a hook add context after a failed tool call \citep{anthropic2026hooks}.
\rev{\revb{Our October 2026 survey found no controlled evaluation of these handlers.} Each documented handler serves its own framework.}
Gateways can intercept tool traffic\revb{,} for example, request and response interceptors in Amazon Bedrock AgentCore \citep{amazonwebservices2026types}\revb{,} but the documented uses are transformation and security, not recovery.
In a survey of documentation and source code in October 2026, we found no inference engine (vLLM, SGLang, llama.cpp, Ollama \citep{ollama2026ollama}) or LLM gateway (LiteLLM \citep{berriai2026litellm}, Gravitee \citep{gravitee2026proxy}) that rewrites or annotates failed tool results; they render tool messages through the chat template as the client sent them.

\paragraph{Context engineering.}
Practitioners have argued that failed actions should stay in the context, so that the model sees the evidence and stops repeating the action, and that the context should be append-only to keep the prefix cache valid \citep{ji2025context}.
Context editing does the opposite for a different reason: it clears old tool results to save tokens near the context limit \citep{anthropic2026context}.
Our forms span both choices: C2 removes the failed exchange, and C3 keeps it and appends.

\paragraph{Determinism and evaluation variance.}
Agent evaluations are noisy: single-run pass@1 on SWE-Bench Verified varies by 2.2 to 6.0 points across runs \citep{bjarnason2026on}, and same-input reruns of agents change their outcomes \citep{yagubyan2026how,bellibatlu2026same}.
Replay harnesses have been proposed to make tool-using agents auditable \citep{khatchadourian2026replayable}.
We use a deterministic engine for a narrower purpose: two delivery forms are compared at the same failure point with the same prefix, so any difference in the next action comes from the form, and every rerun reproduces the same tokens.

%% ===================================================================
\section{\rev{Methods: Error Delivery and Implementation}}
\label{sec:methods}
\subsection{\rev{Problem and Design Space}}
\label{sec:problem}

\subsubsection{\rev{Tool errors in the agent loop}}
An LLM agent alternates between model turns and tool executions.
The model emits a tool call, the client runs it and appends the result as a tool message, and the next request to the inference engine carries the whole history.
When the tool fails, nothing else happens: the failure reaches the model only as text inside that history.
We call the first request that contains a failed tool result the \emph{decision step}.
The model's next action is decided there.
It can call the same tool again, call it with different arguments, call something else, or answer in text.

\subsubsection{\rev{Error kinds and the correct action}}
\label{sec:kinds}
What counts as recovery depends on the error.
We use four kinds (Table~\ref{tab:kinds}).
\rev{A \emph{transient} error, such as a dropped connection or a rate limit, is expected to clear. Repeating the same call is the scored action in our controlled tasks; it is an appropriate operational action only when repetition is safe. A timeout alone does not establish that a side-effecting operation failed to commit.}
A \emph{fixable} error names a value to change, as in ``\texttt{old\_string} must not be empty'', so the correct action is to call again with the corrected value.
An \emph{unrecoverable} error, such as a permission denial, cannot be fixed by the agent, so the correct action is to stop calling and tell the user.
An \emph{ambiguous} error, such as a missing file or a non-zero exit code, has no single correct action: the agent may need another path, another command, or a report.
The first three kinds have a correct next action that a short sentence can state; the fourth does not.

\begin{table}[t]
\caption{Error kinds, the correct next action, and the guidance sentence attached by the engine (verbatim).}
\label{tab:kinds}
\small
\begin{tabular}{@{}lp{3.2cm}p{7.2cm}@{}}
\toprule
Kind & Correct action & Guidance sentence \\
\midrule
Transient & Repeat the same call & \textit{This is a temporary failure. Retrying the same call is expected to succeed.} \\
Fixable & Call again with the value the error names & \textit{This call cannot succeed as written. The error says what to change; call the tool again with the corrected value.} \\
Unrecoverable & Stop and tell the user & \textit{This cannot be fixed by retrying or by changing the arguments. Tell the user what happened.} \\
Ambiguous & Depends on context & (none\revb{,} the result is left unchanged) \\
\bottomrule
\end{tabular}
\end{table}

\subsubsection{\rev{Dimensions of delivery}}
\label{sec:dimensions}
Given a failed result, the text the model sees can vary along four dimensions.
\begin{itemize}
\item \textbf{Placement.} The error stays inside the tool message, where the client put it, or it is restated as a note from the tool runtime in a user turn.
\item \textbf{Guidance.} No guidance, or one kind-specific sentence (Table~\ref{tab:kinds}).
\item \textbf{History.} The failed call and its result stay in the history, or they are removed and replaced by the note.
\item \textbf{Turn structure.} The note is appended to an existing message, or it opens a new user turn after the failed exchange.
\end{itemize}
We name the forms that the studies compare.
C0 is the conversation as the client sent it.
C1 appends the guidance sentence to the tool message.
C2 removes the failed call and its result and appends the note with guidance to the last remaining message if that is a user message, and otherwise adds it as a new user message; it is the form that Study~1 selected (A2B2 in Section~\ref{sec:study1}). In Study~1's single-call scenarios the last remaining message is always the user request; in histories with several calls per message, C2 can produce the same messages as C4 (Table~\ref{tab:study2forms}).
C3 keeps the history unchanged and places the note with guidance in the first user message after the failed exchange, or in a new user message if there is none.
At the decision step, C3 keeps the history in the sense of \citet{ji2025context}: nothing earlier in the conversation is removed. In the engine the note exists only in the rewritten request; the client's own history does not contain it, so later requests do not either.

\begin{table}[t]
\caption{Study~2: the five conditions. The note and guidance text of C2, C3 and C4 is identical.}
\label{tab:study2forms}
\small
\begin{tabular}{@{}llp{4.4cm}l@{}}
\toprule
Condition & Failed call and result & Note and guidance & History kept \\
\midrule
C0 & kept & none & yes \\
C1 & kept & guidance at the end of the tool message & yes \\
C2 (Study~1's A2B2) & removed & at the end of the last message if it is a user message, else in a new user turn & no \\
C3 & kept & in a new user turn after the failure & yes \\
C4 & removed & in a new user turn & no \\
\bottomrule
\end{tabular}
\end{table}

\subsubsection{\rev{Why the inference engine}}
\rev{The engine renders each incoming conversation through the model's chat template, giving this implementation a common point for changing the model-visible history. This avoids modifying each agent separately. It does not establish an engine-specific scientific advantage: a client or gateway that can edit the same messages could implement the same forms. We do not compare equivalent engine-side and proxy-side transformations.}

\rev{This placement has two limits. The engine infers failures from flags and text rather than from the tool's execution state, so detection can be incomplete or wrong. It also changes what the model reads, so rewriting is opt-in and every detection is visible to the client. The intervention changes conversation delivery, not model weights or tool execution.}

%% ===================================================================

\begin{figure}[t]
\centering
\includegraphics[width=\linewidth]{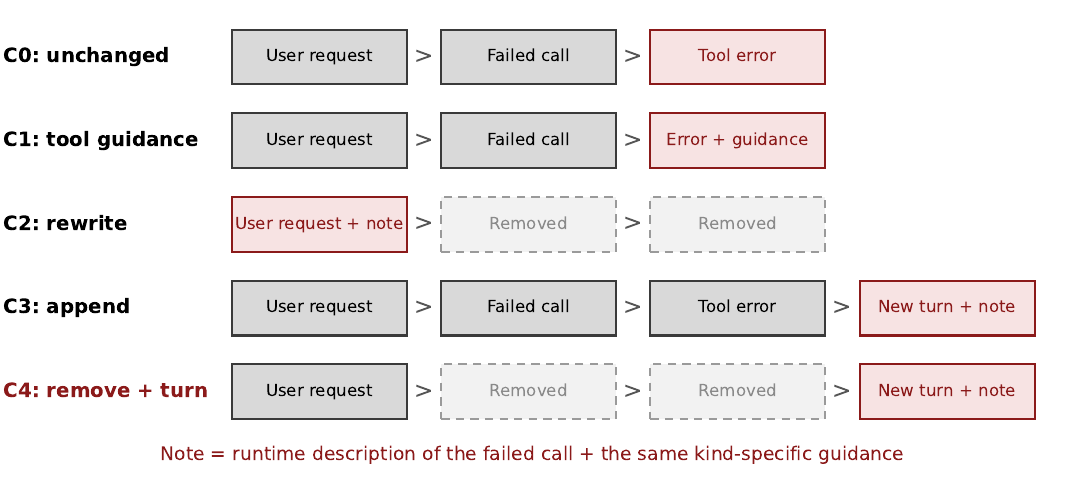}\Description{Five rows of boxes, one per delivery form. C0 (unchanged): user request, failed call, tool error. C1 (tool guidance): user request, failed call, error with guidance. C2 (rewrite): user request with the note; the failed call and the tool error are removed. C3 (append): user request, failed call, tool error, then a new turn with the note. C4 (remove and turn): the failed call and the tool error are removed, then a new turn with the note.}
\caption{\rev{Delivery forms at the decision step. Boxes show conversation structure, not measured outcomes. C1 adds guidance to the tool result; C2 removes the failed exchange and changes the earlier user message (or adds a user message if the last remaining message is not one); C3 keeps the exchange and adds a later user turn. \revb{C4 removes the failed exchange and adds the same note in a new user turn.} The note is runtime-supplied, not a new human instruction.}}
\label{fig:forms}
\end{figure}
\rev{The intervention is a message transformation: it does not run tools or change their results. The following implementation makes that transformation available to existing agents. A chat template is the model-specific format used to serialize conversation messages; a prefix cache reuses computation for text already processed.}

\subsection{\rev{Engine-Level Tool-Error Correction}}
\label{sec:design}

\subsubsection{\rev{Overview}}
The layer runs once per request, after the request has been converted to a common chat representation and before the chat template is rendered.
The engine accepts three API surfaces\revb{,} OpenAI Chat Completions, Anthropic Messages and OpenAI Responses\revb{,} and converts each to OpenAI-style \texttt{messages}; the layer therefore applies to all of them in the same way.
The same rendering path serves generation, \texttt{/tokenize} and token counting, so a client that counts tokens sees the counts of the text that the model actually reads.

The layer has three steps: it detects failed tool results (Section~\ref{sec:detect}), classifies each failure into one of the kinds of Section~\ref{sec:kinds} (Section~\ref{sec:classify}), and, if a form is enabled, rewrites the conversation into that form (Section~\ref{sec:forms}).
Detection is always on and only recorded; rewriting is off by default.

\subsubsection{\rev{Detection and classification}}
\label{sec:detect}
\label{sec:classify}
\rev{Detection asks whether a tool result signals a failure; classification asks what kind of failure it is. The layer first reads explicit client flags, then applies recorded rules for known agents, and finally checks standard error prefixes for other clients. Classification uses the error wording and leaves uncertain cases ambiguous, with no guidance. This conservative design avoids treating every occurrence of an error word as a failed call. The complete detection procedure, classification priorities and examples are retained in Appendix~\ref{app:implementation}; the exact rule lists are in Appendix~\ref{app:rules}.}

\subsubsection{\rev{Forms}}
\label{sec:forms}
When a form is enabled, each detected failure of a non-ambiguous kind is rewritten.
The runtime note reads, verbatim,
\begin{center}\scriptsize
\texttt{Note from the tool runtime: the last attempt called \{name\} with \{arguments\} and it failed with: \{error\}}\\
\texttt{\{guidance\}}
\end{center}
where the arguments are printed as \texttt{key=value} pairs and the guidance is the sentence of Table~\ref{tab:kinds}.
C1 appends a line break and the guidance sentence to the tool message.
C2 removes the failed call and its result and appends two line breaks and the note to the last remaining message if it is a user message; otherwise it adds the note as a new user message.
C3 leaves the history unchanged and appends two line breaks and the note to the first user message after the failed exchange; if no such message exists, it inserts a new user message at that point.
The texts are byte-for-byte those of the registered study code, so that the engine reproduces exactly the conditions that the studies measured (Section~\ref{sec:verify}).

\subsubsection{\rev{Scope and cost}}
Only failures in the tool results after the last assistant message are rewritten, that is, only at the decision step.
Earlier failures are left as the client sent them: the studies measured the decision step, and no study measured a form applied to the whole history.
Because the next request again shows the failure in its original form, the rewrite affects exactly one turn.
Its cost is a partial loss of the prefix cache for that turn: C2 recomputes from the removed call, C3 from the failed result and the inserted note.
In Study~2 the median recomputed prompt at the decision step was 94 tokens for C2 and 171 for C3 on synthetic items, and 965 and 1,081 on real histories (\revblue{Section~\ref{sec:results2}}).

\subsubsection{\rev{Selection and visibility}}
A request field (\texttt{geulpu\_tool\_errors}: \texttt{off}, \texttt{c1}, \texttt{c2} or \texttt{c3}) overrides a server option, which overrides the default (off).
Every detection is written to the server log and to a response header, \texttt{X-GeulPU-Tool-Errors}, which lists each failed message with its kind and the rule that fired, the active form, and whether the conversation was changed; the response body is unchanged.
A client can thus see, request by request, what the model was shown.

\subsubsection{\rev{Implementation and verification}}
\label{sec:verify}
\rev{The layer is implemented in \revb{an open-source deterministic inference engine} \citep{woo2026geulpu}. Studies~1 and~2 use its exact mode: the same request yields identical output tokens on rerun, so paired form comparisons do not include sampling variation. This helps isolate delivery effects but does not test sampled decoding. Engine and precision details are in Appendix~\ref{app:implementation}.}

We verify the layer in three ways.
First, 75 detection cases (results recorded from real agents, standard error texts, and texts that must \emph{not} be detected) give the expected kind and rule, and 49 form cases match a Python reference implementation character for character.
Second, on 270 items built from 30 points in real coding-agent histories (Study~2), the engine's C1, C2 and C3 conversations equal those of the registered study code byte for byte.
Third, in our checks, requests with no detected failure produced the same bytes as the engine without the layer.

% Added 2026-10-31 (S. Woo): ACM's authorship policy asks that AI used to conduct the research be described in the methods section.
\subsection{\revblue{Use of AI Tools}}
\label{sec:ai}
\revblue{AI coding and writing tools were used throughout this research. Under the first author's direction, they were used to write the engine and the tool-error layer, the stimulus generators, the run scripts and the analysis scripts, to operate the measurement runs, and to draft the registrations and this manuscript. The registered analysis scripts were fixed by hashes before the data they analyze existed (Section~\ref{sec:experiments}), the forms were checked against a reference implementation (Section~\ref{sec:verify}), and the reported measurements were taken from the released analysis outputs \citep{woo2026study}. The authors take responsibility for the content.}

%% ===================================================================
\section{\rev{Experiments}}
\label{sec:experiments}
\rev{The experiments build on one another. Study~1 identifies which delivery components change the next action. Study~2 tests whether a new turn can retain the failed exchange without losing those benefits. Study~3 asks whether the resulting implementation helps complete tasks when detection is imperfect. In all studies, the intervention changes the conversation rather than the model weights.}
\subsection{\rev{Study 1: Does delivery change the next action?}}
\label{sec:study1}
\rev{We first separate placement from guidance. A1 leaves the error in the tool message; A2 restates it in the earlier user message and removes the failed exchange. B1 adds no guidance; B2 adds one kind-specific sentence. The four combinations therefore ask whether restating the error helps, whether guidance helps, and whether delivery still matters when guidance is present. A2 changes both placement and retained history, a distinction examined further in Study~2.}

\rev{The main experiment uses 60 synthetic scenarios over six read-only tools, three error kinds and four forms: 720 conversations per model. Six open models run greedily, and a registered extension adds two 32B models. The primary outcome is the next action; episodes then continue for up to five turns to measure completion and futile calls. A fixed selection rule balances correct recovery against correct stopping. The design and analysis were fixed before data collection, with local hash records. \revb{The exact tests and thresholds of these hypotheses} and the model settings are retained in Appendix~\ref{app:study1}; the full follow-up results are in Appendix~\ref{app:results1}, with the main findings in Section~\ref{sec:results1}.}

\revb{The five hypotheses, in plain terms, are the following.}
\begin{itemize}
\item[\textbf{H1.1.}] Moving the error into a runtime note, without guidance, raises the share of correct next actions on transient and fixable errors (A2B1 against A1B1; supported if positive at $p<0.05$ in at least four of the six models).
\item[\textbf{H1.2.}] Adding one guidance sentence inside the tool message raises that share (A1B2 against A1B1; same rule).
\item[\textbf{H1.3.}] With guidance present, placement still changes the outcome (A2B2 against A1B2; decided if at least four models differ in the same direction).
\item[\textbf{H1.4.}] No form lowers correct stopping on unrecoverable errors by more than 10 points against the client's form.
\item[\textbf{H1.5.}] The selected form beats the client's form over whole episodes: higher completion in at least four models, and no more than $+0.5$ added futile calls, pooled.
\end{itemize}

\subsection{\rev{Study 2: Can a new turn preserve the useful evidence?}}
\label{sec:study2}
\rev{Study~1 selected C2, but its follow-ups exposed a stopping failure when the failed exchange was removed. Study~2 therefore tests C3: keep the exchange and place the same note and guidance in a later user turn. C4 removes the exchange but also creates a new turn, helping separate retained history from note position. The forms are defined in Table~\ref{tab:study2forms}.}

\rev{Fresh synthetic scenarios give 900 conversations per model. The small tier has seven models and the large tier six. A second component uses 30 failure points from six real Qwen Code sessions, with three injected error kinds and C0--C3: 360 items, grouped into 19 assistant-message clusters. The primary tests ask whether C3 is non-inferior to C2 on synthetic errors and superior on real histories; secondary tests examine removal, placement, stopping, episode outcomes and cache cost. The small-tier plan was registered on OSF before data existed; the large-tier extension was fixed locally without an external timestamp. All registered thresholds, adjustments, model details and deviations are in Appendix~\ref{app:study2}.}

\revb{The two primary hypotheses, decided together by Holm's procedure at $\alpha = 0.05$, are the following.}
\begin{itemize}
\item[\textbf{H2.1.}] C3 is not worse than C2 on synthetic errors, pooled over models and error kinds, within a 5-point non-inferiority margin.
\item[\textbf{H2.2.}] On real coding-agent histories, C3 gives more correct actions than C2.
\end{itemize}
\revb{Secondary checks: S1, removing the failed exchange neither helps nor hurts (C4 against C3 within $\pm 5$ points); S2, guidance in a new turn beats guidance in the tool message (C3 above C1 in at least five of the seven models); S3, C2 against C4 (descriptive); S4, for Llama-3.1, C3 restores stopping on unrecoverable errors where C2 does not stop; S5, over whole episodes, C3 does not lose completion ($-5$ points) or add futile calls ($+0.5$); S6, the prompt tokens recomputed at the decision step (descriptive).}

\subsection{\rev{Study 3: Does the implemented layer help complete tasks?}}
\label{sec:study3}
\rev{The first two studies intervene at a known failure point. Study~3 adds a practical constraint: the shipped layer must detect and classify errors itself. \revb{G-off and G-on are the same deterministic inference engine with the recovery layer disabled and enabled: with G-off, the agent receives the original tool-error feedback without the layer's intervention, and with G-on, the layer applies the C3 form (Table~\ref{tab:study2forms}). The within-engine comparison therefore isolates the effect of enabling the layer, without changing engines.} \revblue{It was not one of the registered hypotheses below; we report it descriptively, with 95\% task-bootstrap intervals.} llama.cpp, vLLM and SGLang provide cross-engine comparisons, while two client alternatives provide exact-kind guidance (E1) or tool-side retries (E2). E1 is an oracle for the injected fault kind, not a matched proxy implementation of C3.}

\rev{Layer~1 uses Loud/Quiet's 24 multi-turn tasks and ToolMaze's 100 tasks, with the first 50 ToolMaze tasks per mode in the large tier. The outcome is final-state agreement for Loud/Quiet and pass rate for ToolMaze. Within-engine G-on versus G-off isolates the implementation more directly; cross-engine comparisons also require baseline checks because quantization, templates and tool-call handling differ. Rules were frozen before ToolMaze was examined.}

\revb{Each tier and benchmark is analyzed separately, with a bootstrap over tasks. The hypotheses are the following.}
\begin{itemize}
\item[\textbf{H3.1.}] On explicit errors (Loud/Quiet's timeout, schema drift and missing tool; ToolMaze's P1 and P2), G-on beats each of llama.cpp, vLLM and SGLang.
\item[\textbf{H3.2.}] G-off starts from the same baseline as each engine, within $\pm 5$ points over all conditions; otherwise a G-on gain is reported as mixed with engine differences.
\item[\textbf{H3.3.}] G-on against the client alternatives E1 and E2 (descriptive).
\item[\textbf{H3.4.}] On Loud/Quiet's missing tool, G-on stops calling the hidden tool more often than each engine (descriptive).
\end{itemize}

\rev{Layer~2 runs Qwen Code on six repositories with Qwen3-8B and Qwen3-32B. Each task fetches data, writes a result file and checks it with tests. The service either works, fails transiently, or refuses access. The last condition tests whether the agent reports a barrier instead of fabricating a result. With one agent and six tasks, these results are descriptive. The pre-measurement plan was fixed in local hash logs after pilots, without a public timestamp. Full engine settings, \revb{the formal versions of these hypotheses}, scoring rules, controls and all six deviations are retained in Appendix~\ref{app:study3}.}

\section{\rev{Results}}
\label{sec:results}

\subsection{\rev{Study 1: Placement and Guidance}}
\label{sec:results1}

\rev{Placement and guidance both improve the next action for most models, but the selected history-removing form fails to stop Llama-3.1-8B. Figure~\ref{fig:study1} makes this tradeoff visible; \rev{Table~\ref{tab:study1} in the appendix retains every cell.}}
In the client's form (A1B1), five of the six models rarely repeated a call after a transient error (0--37\%) and corrected a fixable one in only 2--80\% of cases; Llama-3.1-8B was the exception at 98\%.
Moving the error into a runtime note (A2) made the same five models act correctly almost always.

\begin{figure}[t]
\centering
\includegraphics[width=\linewidth]{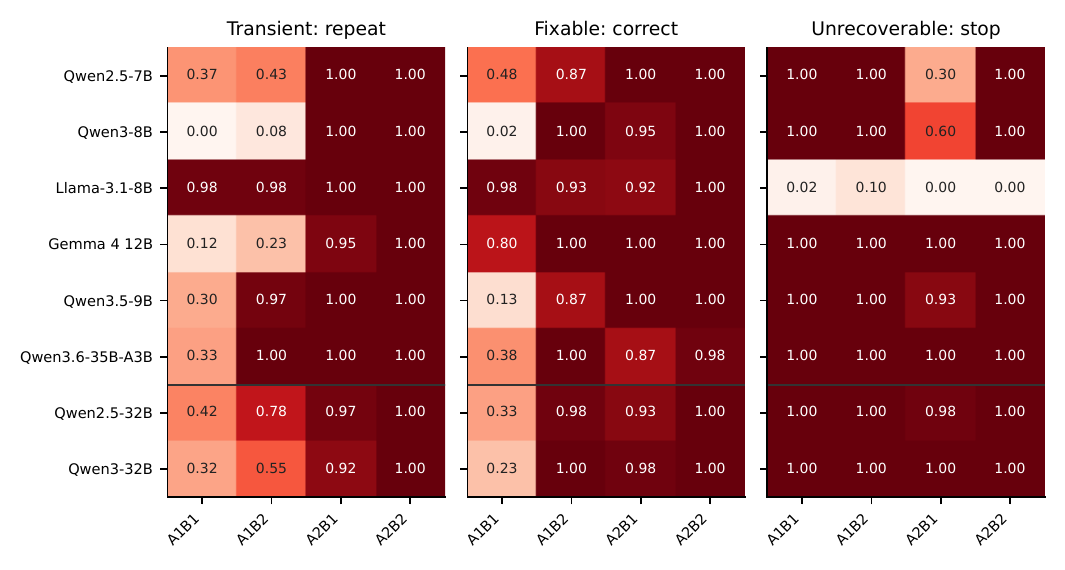}\Description{Three heatmaps, one per error kind (transient: repeat; fixable: correct; unrecoverable: stop). Rows are eight models and columns are the forms A1B1, A1B2, A2B1 and A2B2; each cell gives the share of correct next actions.}
\caption{\rev{Study 1 decision-step accuracy, reproduced from Table~\ref{tab:study1}. Each cell contains the reported share; no uncertainty is inferred. Moving the error and adding guidance works well for transient and fixable errors, but Llama-3.1-8B does not stop under either A2 condition. The final two rows are the registered larger-model extension.}}
\label{fig:study1}
\end{figure}

\paragraph{\rev{Placement and guidance help, but are not interchangeable.}}
\rev{\revb{H1.1} (placement without guidance) and \revb{H1.2} (guidance in the tool message) were each supported in five of six models. Even with guidance held fixed, delivery still mattered: A2B2 outperformed A1B2 in four models, with no significant difference in the opposite direction (\revb{H1.3}). Without guidance, however, moving the error made stopping worse for two models. The result is not simply ``move the error'': the delivery form must preserve correct stopping as well as encourage recovery.}

\paragraph{\rev{The selected form exposes a stopping failure.}}
\rev{The registered rule selected A2B2, implemented as C2, because it combined strong recovery with the pooled stopping criterion. Yet the whole-episode test \revb{H1.5} was not supported: futile calls increased beyond its margin. This came from Llama-3.1-8B, which kept calling after unrecoverable errors. The larger-model extension confirmed the recovery benefit at 32B, but \revblue{included no Llama model, so it could not test that failure}.}

\paragraph{\rev{The follow-up points to turn structure.}}
\rev{Stronger wording did not fix the Llama failure: even ``Do not call any tool again.'' left stopping at 0 of 60. A new user message after the failed exchange restored stopping in 59 of 60. Keeping the failed exchange also mattered. These interventions support a structural explanation for this setting; they do not establish an internal cognitive mechanism. They motivate C3, which keeps the failed exchange and adds a later user turn. Exploratory real-session checks also warned that removing history could lose useful context. All exact hypothesis statistics, extensions, rerun checks and follow-up outcomes are retained in Appendix~\ref{app:results1}.}

\subsection{\rev{Study 2: Retaining History and Adding a Turn}}
\label{sec:results2}

\begin{figure}[H]
\centering
\includegraphics[width=\linewidth]{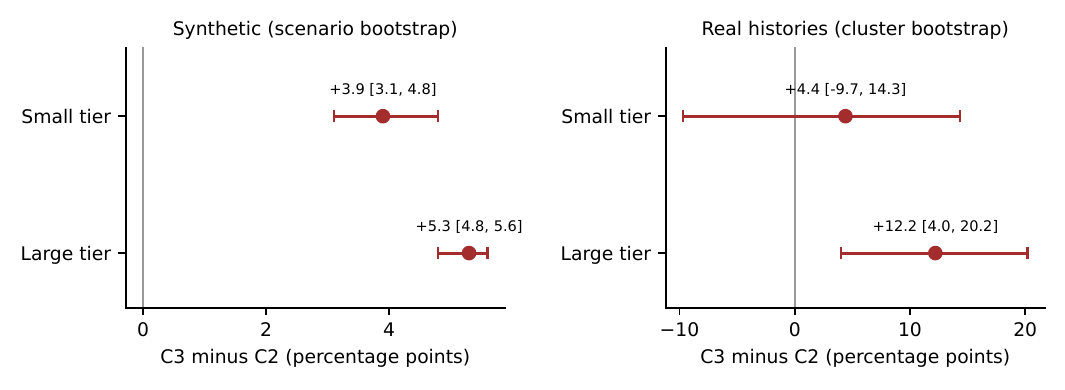}\Description{Two panels of point estimates with 95\% intervals for C3 minus C2, in percentage points, for the small and large tiers. Synthetic errors: +3.9 [3.1, 4.8] and +5.3 [4.8, 5.6]. Real histories: +4.4 [-9.7, 14.3] and +12.2 [4.0, 20.2].}
\caption{\rev{Study 2: C3 minus C2 in percentage points. Left: pooled synthetic decision-step accuracy with reported 95\% scenario-bootstrap intervals. Right: real-history accuracy with reported 95\% cluster-bootstrap intervals over 19 assistant messages. The real histories contain 30 failure points; limited clusters constrain generalization. These panels use different resampling units and must not be pooled.}}
\label{fig:study2}
\end{figure}

\rev{C3 passes the pooled synthetic non-inferiority test in both tiers and avoids the Llama stopping failure. Its real-history advantage is supported only in the large tier (Figure~\ref{fig:study2}; Table~\ref{tab:study2}). These are distinct findings, not evidence of a universal optimum.}
No model exceeded the registered 20\% limit for format errors (the highest was 5.2\%, Llama-3.2-3B), and the reruns were exact: 630 of 630 synthetic and 36 of 36 real conversations reproduced every turn's tokens, and Study~1's 540 Qwen3-8B conversations, rerun through Study~2's code, gave the same tokens as before.

\begin{table}[t]
\caption{Study~2: registered results. $\Delta$ values are percentage points with 95\% bootstrap intervals (S1: 90\%). Correct action on real histories: share of 90 items per condition, on one model per tier (small: Qwen3.6-35B-A3B; large: Qwen3.6-27B).}
\label{tab:study2}
\footnotesize
\begin{tabular}{@{}p{3.9cm}p{4.6cm}p{4.6cm}@{}}
\toprule
& \rev{Small tier (7 models, 3--35B)} & Large tier (6 models, 27--70B) \\
\midrule
H2.1: C3 $-$ C2, synthetic & $+3.9$ $[3.1, 4.8]$, supported & $+5.3$ $[4.8, 5.6]$, supported \\
H2.2: real histories, C3 vs.\ C2 & 55.6\% vs.\ 51.1\%, $p = 0.58$, \mbox{not supported} & 57.8\% vs.\ 45.6\%, $p = 0.027$, \mbox{supported} \\
\quad C3 $-$ C0, real histories & $+34.4$ $[25.6, 43.3]$ & $+34.4$ $[24.4, 44.4]$ \\
\quad C3 $-$ C2, cluster bootstrap & $+4.4$ $[-9.7, 14.3]$ & $+12.2$ $[4.0, 20.2]$ \\
S1: C4 $-$ C3 & $-3.8$ $[-4.5, -3.2]$, equivalent & $-1.9$ $[-2.5, -1.2]$, equivalent \\
S2: C3 $>$ C1 & 5 of 7 models, supported & 5 of 6 models, supported \\
S4: Llama stopping, C3 $-$ C2 & $+98.3$ $[95, 100]$, supported & $+98.3$ $[95, 100]$, supported \\
S5: completion, C3 $-$ C2 & $+4.2$ $[1.9, 6.4]$, supported & $+0.4$ $[-1.9, 2.5]$, supported \\
S5: futile calls, C3 $-$ C2 & $-0.71$ $[-0.74, -0.66]$, supported & $-0.80$ $[-0.83, -0.75]$, supported \\
\bottomrule
\end{tabular}
\end{table}

\paragraph{\rev{The pooled benefit is driven largely by Llama.}}
\rev{C3 exceeds C2 on the pooled synthetic comparisons in both tiers (Figure~\ref{fig:study2}), passing the registered non-inferiority test H2.1. The largest gains come from avoiding the Llama stopping failure: for Llama-3.3-70B, stopping on unrecoverable errors is 95\% with the client's form, 2\% with C2 and 100\% with C3, so C2 itself breaks a stopping behavior that works without intervention. Without Llama-3.1-8B and Llama-3.3-70B, the pooled difference is $-0.1$ points in the small tier and $-0.2$ in the large tier. Other models gain little, and Qwen2.5-7B is worse with C3. C3 is therefore an aggregate choice with model-dependent tradeoffs, not a universal optimum.}

\paragraph{\rev{Real histories give a more qualified answer.}}
\rev{The real-history superiority test H2.2 is not supported in the small tier (55.6\% versus 51.1\%, $p=0.58$). It is supported in the large tier (57.8\% versus 45.6\%, $p=0.027$). The cluster-bootstrap intervals in Figure~\ref{fig:study2} retain the dependency among failure points within the same assistant message. With only 19 clusters from six sessions, these results have a limited basis for generalization. On these histories C2 adds a new user turn, as C4 does, so C3 against C2 contrasts retained and removed history rather than turn position. Exploratory sensitivity analyses (Appendix~\ref{app:results2}) agree in direction but leave a narrow margin: all six sessions favor C3 in the large tier and permutation tests over messages or sessions give $p \approx 0.03$, but leaving out one session raises $p$ to as much as 0.077, and if responses that add other calls to the correct one are not counted as correct, the large-tier difference shrinks to 6.7 points ($p = 0.21$).}

\paragraph{\rev{The benefit is not a cache saving at the measured step.}}
\rev{C3 reduces futile calls relative to C2 and meets the episode-completion criterion, but recomputes more prompt tokens at the decision step: medians are 171 versus 94 on synthetic items and 1,081 versus 965 on real histories. Any later-turn cache benefit remains unmeasured. Table~\ref{tab:study2} retains the registered outcomes; Appendix~\ref{app:results2} retains every model-specific comparison, rerun check and exploratory corrective-gain result.}

\subsection{\rev{Study 3: Public Benchmarks}}
\label{sec:results3}

\rev{The within-engine comparison isolates the implemented layer more directly than comparisons across engines: ToolMaze improves, while Loud/Quiet shows no measurable benefit (Figure~\ref{fig:study3}). \rev{Table~\ref{tab:study3l1} in the appendix retains the cross-engine baseline checks and client alternatives.}}

\begin{figure}[H]
\centering
\includegraphics[width=\linewidth]{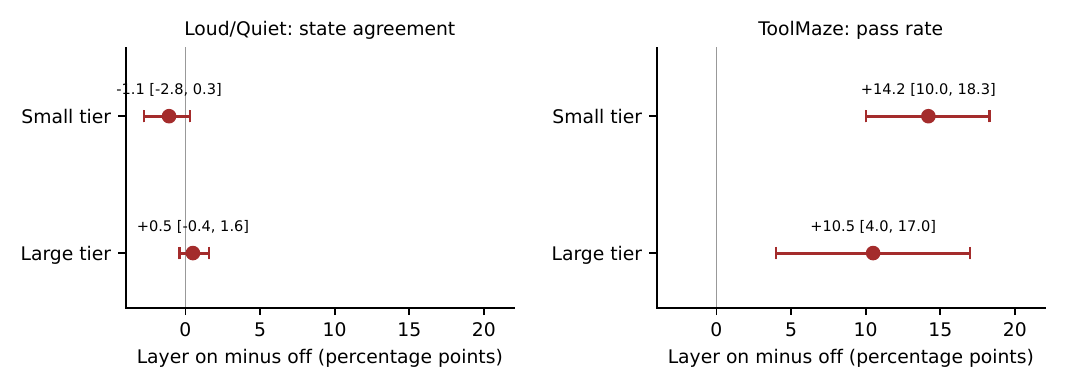}\Description{Two panels of point estimates with 95\% intervals for the layer on minus off on explicit errors, in percentage points, for the small and large tiers. Loud/Quiet state agreement: -1.1 [-2.8, 0.3] and +0.5 [-0.4, 1.6]. ToolMaze pass rate: +14.2 [10.0, 18.3] and +10.5 [4.0, 17.0].}
\caption{\rev{Study 3: G-on minus G-off on explicit errors, in percentage points, with reported 95\% task-bootstrap intervals from Table~\ref{tab:study3l1}. Loud/Quiet measures final-state agreement; ToolMaze measures pass rate. A common percentage-point axis does not make the outcome measures interchangeable. Positive values favor the layer.}}
\label{fig:study3}
\end{figure}

\paragraph{ToolMaze.}
The layer raised the pass rate on explicit errors by 14.2 points in the small tier and by 10.5 in the large tier, \rev{with no statistically detectable change where there was no fault (P0).}
\revb{H3.1} was supported against all three engines in the small tier, and because G-off's baseline equalled those of vLLM and SGLang (\revb{H3.2}), \rev{the results support a contribution from the layer, while not eliminating all cross-engine differences.}
The gain came mostly from transient faults: Qwen3-8B passed P1 in 3\% of runs with G-off and in 43\% with G-on, and Qwen2.5-7B went from 0\% to 31\% while every other engine stayed at 0\%.
In the large tier, \revb{H3.1} was supported against llama.cpp and SGLang but not vLLM ($p = 0.052$); the large gain over llama.cpp reflects a baseline gap, since llama.cpp passed only 24\% of fault-free Llama-3.3-70B runs against \revb{G-off's} 82\%.
Llama-3.3-70B gained most on permanent faults (28\% to 48\%).
The oracle client E1 beat the layer by 16.8 and 18.5 points.
\rev{This is consistent with a coverage limit, but is not a matched-detector ablation:} E1 knows the kind of every injected fault, whereas the frozen rules give the correct guidance to only 29.8\% of ToolMaze's explicit errors.

\paragraph{Loud/Quiet.}
The layer had no effect ($-1.1$ and $+0.5$ points), and E1 and E2 did not differ from it.
\revb{H3.1} was nominally supported in the small tier, but \revb{H3.2} shows why: llama.cpp, vLLM and SGLang ended 137, 134 and 120 runs with errors, all with Llama-3.1-8B\revb{, whose} tool calls llama.cpp could not parse \revb{and whose} multi-call histories vLLM and SGLang rejected with HTTP~400\revb{,} so their state agreement for that model was 0.34--0.41 against \revb{G-off's} \revblue{0.81}--0.85.
For Qwen3-8B alone, G-on was 0.6 to 2.1 points \emph{below} the other engines.
\revb{H3.4} was not supported: the missing tool is classified as ambiguous, so the layer gives no guidance there, and the differences follow the engines' baselines.

\subsection{\rev{Study 3: Coding-Agent Tasks}}
\rev{Table~\ref{tab:study3l2} in the appendix gives layer~2} with the \revb{engine's} rows re-run under ASCII folder names; in those 72 runs, one tool call broke off.

\begin{figure}[H]
\centering
\includegraphics[width=\linewidth]{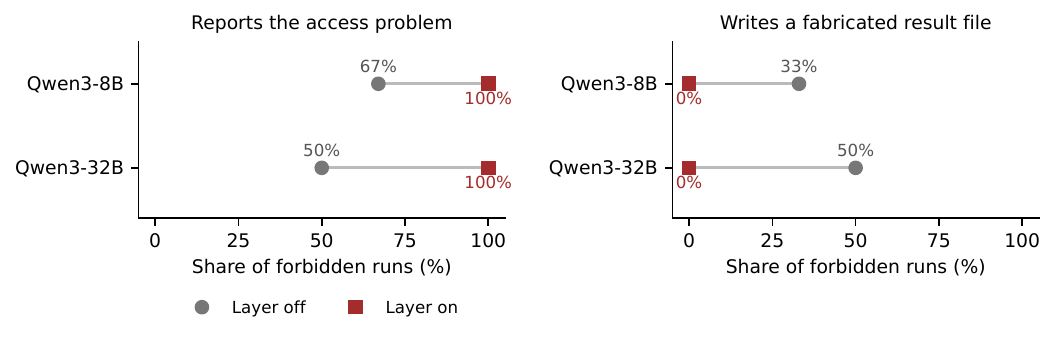}\Description{Two dot plots of the share of forbidden runs for Qwen3-8B and Qwen3-32B, with the layer off (circles) and on (squares). Reports the access problem: 67\% and 50\% with the layer off, 100\% with it on. Writes a fabricated result file: 33\% and 50\% with the layer off, 0\% with it on.}
\caption{\rev{Coding-agent behavior when access is forbidden. Points reproduce the reported model-specific percentages; each model has six tasks per condition. Layer-on reports the access problem in all 12 pooled runs and writes no result file. No uncertainty is inferred from these small descriptive samples.}}
\label{fig:coding}
\end{figure}

\rev{The clearest descriptive effect was on the unrecoverable error (Figure~\ref{fig:coding}).}
With G-on, the agent reported the access problem in all 12 forbidden runs, made a single request to the data service, and never wrote a result file.
With G-off it reported the problem in 67\% (Qwen3-8B) and 50\% (Qwen3-32B) of runs, made 7.7 and 3.8 requests, and wrote a result file\revb{,} necessarily fabricated\revb{,} in 33\% and 50\% of runs; llama.cpp and vLLM made 2.2--8.8 requests and fabricated in 17--33\%.
On the transient error, G-on passed the tests 17 points more often than G-off, but the effect depended on the model: Qwen3-32B passed every flaky run with G-on against half with G-off, while Qwen3-8B passed fewer (33\% against 50\%); vLLM did as well as G-on.
Without a fault, G-on and G-off did not differ ($-8$ points, interval $[-50, 25]$), and G-off's baseline did not differ detectably from llama.cpp, vLLM or the added SGLang run, though with six tasks the intervals span 50 to 100 points.
The confounded \revb{G-on and G-off} rows of the main run, in which the folder names broke the agent's file paths, gave a lower pass rate for G-on without a fault ($-67$ points against G-off); that difference came from the folder names and disappeared under ASCII names.

%% ===================================================================

\section{Discussion}
\label{sec:discussion}

\paragraph{\rev{Delivery effects and implementation choice.}}
The engine layer changed what agents did without any change to the agents.
On ToolMaze it raised success on explicit errors by 10--14 points, with a real coding agent it turned an unrecoverable error into a report to the user in every run, and where nothing failed it did no measurable harm.
Yet a client that knew each injected fault (E1) did better on ToolMaze by 17--19 points.
E1 placed its guidance in the tool message, a form that Studies~1 and~2 found weaker than a note in a user turn (\revb{H1.3}; S2 in five of seven and five of six models), \rev{so its advantage is consistent with better fault information overcoming a weaker delivery form. E1 and G-on also differ in coverage and form; this comparison does not isolate detection as the only cause.}
\rev{This motivates, but does not establish, a division of labor:} the client, which knows what its tool did, marks the failure and its kind\revb{, for which} \texttt{is\_error} and \texttt{isError} already exist\revb{,} and the engine, which renders every agent's history, delivers it in the form the studies support.
We have not measured this combination.

\paragraph{\rev{Detection coverage limits the measured gain.}}
The rules were written from the error texts of languages, libraries and protocols and frozen before the evaluation.
On ToolMaze, whose texts we had not seen, they gave the correct guidance to 29.8\% of explicit errors, the wrong guidance to 3.9\%, and none to 66.3\%; the measured gain is what this coverage buys.
Wrong guidance is the larger risk: Study~1 showed that a note without guidance can make a model call again where it should stop, and a permanent error read as transient invites futile retries.
Ambiguous errors\revb{, such as} a missing file \revb{or} a non-zero exit code\revb{,} receive no guidance by design, which is why the layer had no effect on Loud/Quiet's missing tool.

\paragraph{Where the layer did not help.}
On Loud/Quiet the layer changed nothing, although the rules classified its timeouts and schema drift.
Its measure is the final state of a whole multi-turn task, while the layer acts on one decision step; a single corrected step may not move the final state.
The effects also depended on the model.
With C3, Qwen2.5-7B was 2.8 points worse than with C2 in Study~2, Llama-3.1-8B repeated a transiently failed call in 82\% of cases against 100\% with C2, and in layer~2 Qwen3-8B passed fewer flaky runs with the layer than without it (33\% against 50\%).
\rev{A form selected for aggregate performance is not best for every model. Model-specific form selection is a plausible alternative, but would need a separate, held-out evaluation rather than selection on these same outcomes.}

\paragraph{\rev{Engine robustness is a separate source of failure.}}
\rev{Cross-engine gaps cannot all be credited to recovery. Parsers rejected or lost tool calls, request validators rejected histories or token limits, and template rendering differed. \revb{The engine} also had its own defects, including doubled-brace parsing and a special token that could interrupt a tool call. The affected rows, reruns and later fix are reported in Appendix~\ref{app:robustness}; Study~3 uses the build before that later fix. Baseline checks and explicit failure counts are necessary to distinguish delivery benefits from engine robustness.}

\paragraph{Retrying and side effects.}
The transient guidance tells the model that repeating the call is expected to succeed.
For a tool with side effects, such as a write or a payment, a repeated call may duplicate the action \citep{li2026where}.
The engine cannot know whether a tool is idempotent; our studies used read-only tools and benchmark tools and did not measure side effects.
\rev{The present implementation has no verified idempotency gate. Its retry guidance should therefore be restricted to read-only calls, calls known to be idempotent, or calls protected by an execution system's deduplication guarantee. For other tools, a client must verify the prior operation's outcome before retrying; error wording is not enough. A client-supplied safety flag and withholding of retry guidance when safety is unknown are design requirements for a broader deployment, not features evaluated here.}

\paragraph{\revb{Limitations.}}
All runs used greedy decoding; determinism makes each comparison exact but leaves sampled decoding unmeasured.
Comparisons across engines are confounded by weights (Q8\_0 against BF16 or FP8) and by template rendering, which \revb{H3.2} checks but cannot remove; comparisons within \revb{the same engine} (G-on against G-off, and the forms of Studies~1 and~2) are not.
The outcome measures are the benchmarks' own (state agreement and a rule-based judge); a ToolMaze pass under a permanent fault means that the model stopped, not that the task was completed. Layer~2 detects a report to the user by keywords, which we checked by reading every final message of the forbidden runs (Appendix~\ref{app:study3}), and Study~2's real-history score counts a response as correct if any of its calls is the correct action (Appendix~\ref{app:results2}).
The scenarios of Studies~1 and~2 are synthetic, \rev{layer~2 has one agent and six tasks, and repeated deterministic runs of an identical prompt would not add independent evidence. Broader validation needs new tasks and repositories and another agent, rather than more identical reruns.} \rev{The models come} from three families, all texts are in English, and the guidance sentences are fixed.
Only Study~2's small tier was registered with a public registry; the other registrations are local hash logs, which fix the order of plan and data but give no third-party time stamp.
\rev{Study~3\textquotesingle s six deviations are listed in Appendix~\ref{app:study3}.}
Because C3 restates the tool's arguments and error text in a user turn, text produced by a tool reaches the model in a role with more authority; an error message that carries instructions could exploit this, a risk addressed by work on instruction hierarchies \citep{wallace2024instruction}, which we did not evaluate. Finally, the layer was built and evaluated in our own engine; the forms are byte-identical to the studies' code, but another engine could implement detection differently.

\begin{table}[H]
\caption{\revb{Summary of the main findings. Rows are the registered comparisons of Sections~\ref{sec:experiments} and~\ref{sec:results}, except the descriptive \revblue{G-on vs.\ G-off rows and} coding-agent row.}}
\label{tab:findings}
\small
\begin{tabular}{@{}p{3.2cm}p{5.4cm}p{4.2cm}@{}}
\toprule
Comparison & Main finding & Qualification \\
\midrule
Study 1: note vs.\ tool message, guidance vs.\ none (H1.1, H1.2) & Each improves the next action in five of six models & Without guidance, the note reduces correct stopping for two models \\
Study 1: selected form C2 over whole episodes (H1.5) & Completion higher in five models & Not supported: futile calls exceed the margin, all from Llama-3.1-8B \\
Study 2: C3 vs.\ C2, synthetic errors (H2.1) & $+3.9$ and $+5.3$ points in the two tiers; non-inferiority passed & Gains driven by Llama stopping; Qwen2.5-7B is worse with C3 \\
Study 2: C3 vs.\ C2, real histories (H2.2) & Supported in the large tier (57.8\% vs.\ 45.6\%) & Not supported in the small tier ($p = 0.58$); 19 clusters limit generalization; significant only under the registered any-call scoring \\
Study 3: G-on vs.\ G-off, ToolMaze \revblue{(descriptive)} & $+14.2$ and $+10.5$ points on explicit errors & No detectable change without faults; an oracle client does 17--19 points better \\
Study 3: G-on vs.\ G-off, Loud/Quiet \revblue{(descriptive)} & No measurable effect ($-1.1$ and $+0.5$ points) & The missing tool is ambiguous by design, so the layer gives no guidance \\
Study 3, layer 2: coding agent & Reports the access problem in all 12 forbidden runs with the layer & Descriptive; the transient-error benefit depends on the model \\
\bottomrule
\end{tabular}
\end{table}

\section{Conclusion}
\rev{The same tool-error feedback can produce different recovery behavior when its placement, retained history or turn structure changes. Across three studies with 13 open models, a runtime note with guidance improves many decision-step outcomes. Retaining the failed exchange and adding a new user turn avoids a severe Llama stopping failure, but does not dominate every model or condition.}

\rev{The \revb{implemented layer} improves ToolMaze outcomes on explicit errors and reporting of access failures in a small coding-agent evaluation, while Loud/Quiet shows no measurable benefit. Better-informed client guidance performs better on ToolMaze, indicating that reliable fault information remains important. The evidence supports delivery as an experimental factor and a practical intervention; it does not establish that an inference engine is the only or best place to implement it. Future work should test matched client or proxy implementations, more independent coding tasks, later turns, sampled decoding, and explicit retry-safety controls for tools with side effects.}

\begin{acks}
\revb{This work was supported by the Agencia Estatal de Investigaci\'{o}n (AEI) (doi:\url{10.13039/501100011033}) under Grant ANNA (PID2025-169490OB-C21), by TUCAN6-CM (TEC-2024/COM460) funded by CM (ORDEN 5696/2024) and by European Union's Horizon Europe research and innovation programme under project BRIDGE-AI (Grant 101299050).}
% Pedro's v1 sentence, replaced on 2026-10-31 by the line below (see Section 3.3): \rev{AI coding and writing tools} were used to develop the engine, to run the experiments and to draft this manuscript; the authors checked every number against the released measurement records.
\revblue{The use of AI tools is described in Section~\ref{sec:ai}.}
\end{acks}

\makeatletter
\let\originalbibitem\bibitem
\renewcommand{\bibitem}[2][]{\color{black}\originalbibitem[#1]{#2}\def\thiskey{#2}\def\newkey{kim2026rein}\ifx\thiskey\newkey\color{black}\fi}
\makeatother
\bibliographystyle{ACM-Reference-Format}
\bibliography{references}

\clearpage\appendix
\section{Detection Rules and Sources}
\label{app:rules}

\sloppy
This appendix lists the detection and classification rules of the engine at the commit used in Study~3.
Matching is on the raw text of a tool message; ``starts with'' ignores leading white space only for the generic rules; words are matched as substrings, and some omit their first letter (\texttt{ermission denied}) so that both capitalizations match.

\paragraph{Explicit flags (all clients).}
Rule~1 (\texttt{is\_error}): the tool message carries \texttt{is\_error: true}, or an Anthropic \texttt{tool\_result} with \texttt{is\_error} was converted into it.
Rule~2 (\texttt{cline}): the result is Cline's JSON object (beginning \texttt{\{"query":}) with \texttt{"success":false}.
Rule~3 (\texttt{opencode}): the result begins \texttt{\{"error":\{"type":"tool.execution"}, as OpenCode reports a failed tool.
Rule~14 (\texttt{mcp-iserror}): the result is a JSON object containing \texttt{"isError":true}, the MCP \texttt{CallToolResult} flag \citep{modelcontextprotocol2026tools}.

\paragraph{Agent rules (chosen by User-Agent).}
Qwen Code (\texttt{QwenCode/}): rule~4 (\texttt{qwen-notfound}), the result begins \texttt{File not found:}; rule~5 (\texttt{qwen-exit}), a shell result beginning \texttt{Command:} whose last \texttt{Exit Code:} is a non-zero integer or whose last \texttt{Error:} line is not \texttt{(none)}; rule~6 (\texttt{qwen-error}), the result begins \texttt{Error}, or begins \texttt{File} and contains \texttt{has not been read}.
Codex (\texttt{codex}): rule~7 (\texttt{codex-exit}), \texttt{Process exited with code N} or a result beginning \texttt{Exit code: N} with $N \neq 0$; rule~8 (\texttt{codex-timeout}), \texttt{command timed out}; rule~9 (\texttt{codex-patch}), \texttt{apply\_patch verification failed}.
Kilo Code (\texttt{Kilo-Code/}): rule~10 (\texttt{kilo-notfound}), the result begins \texttt{File not found:}; rule~11 (\texttt{kilo-fs}), it begins \texttt{Unknown: FileSystem.}; rule~12 (\texttt{kilo-usage}), it begins \texttt{oldString}.
OpenCode (\texttt{opencode/}): rule~13 (\texttt{opencode-exit}), the last line \texttt{Exited with code N} has $N \neq 0$.
These rules were recorded from real sessions of each agent; for these agents the words inside shell output are not widened, because a script's own error (for example a \texttt{TypeError} printed by a test) is not a failure of the tool call.

\paragraph{Generic rule (rule~15, unknown User-Agent).}
A result is a failure if it starts with one of the following texts (Table~\ref{tab:generic}).
A JSON \texttt{error} field that is \texttt{null}, \texttt{false} or an empty string is not a failure.

\begin{table}[h]
\caption{Start texts of the generic rule and their sources.}
\label{tab:generic}
\footnotesize
\begin{tabular}{@{}p{6.4cm}p{6.6cm}@{}}
\toprule
Start text & Source \\
\midrule
\texttt{Error:}, \texttt{ERROR:}, \texttt{error:} & LangChain and LangGraph tool nodes; common CLI form \\
\texttt{Error executing tool} & MCP Python SDK (FastMCP) \\
\texttt{MCP error -} & MCP TypeScript SDK (\texttt{McpError}) \\
\texttt{An error occurred while running the tool} & OpenAI Agents SDK default failure function \\
\texttt{Traceback (most recent call last):} & Python exception trace \\
an exception name ending in \texttt{Error}, \texttt{Exception}, \texttt{Timeout} or \texttt{timeout}, followed by \texttt{:} or \texttt{(} & Python exception string form (\texttt{TypeError: \ldots}, \texttt{ConnectionResetError(\ldots)}) \\
\texttt{HTTP Error }, \texttt{HTTP 4}, \texttt{HTTP 5} & \texttt{urllib} \texttt{HTTPError}; HTTP status lines \\
\texttt{NNN Client Error:}, \texttt{NNN Server Error:} & \texttt{requests} \texttt{raise\_for\_status} \\
\texttt{Client error '}, \texttt{Server error '} & \texttt{httpx} \texttt{raise\_for\_status} \\
\texttt{\{"error":} followed by an object or a non-empty string & common API error bodies \\
\bottomrule
\end{tabular}
\end{table}

\paragraph{Kinds.}
The kind is decided from the first 4\,KB of the error text, in the order unrecoverable, transient, fixable; anything else is ambiguous and gets no guidance.
For every rule the common words apply.
Unrecoverable: \texttt{EACCES}, \texttt{EPERM}, \texttt{ermission denied}, \texttt{Access is denied}, \texttt{ead-only}, \texttt{Unable to write}, \texttt{Forbidden}.
Transient: \texttt{ECONNRESET}, \texttt{ECONNREFUSED}, \texttt{ETIMEDOUT}, \texttt{EAI\_AGAIN}, \texttt{Too Many Requests}, \texttt{ate limit}, \texttt{Service Unavailable}, \texttt{Gateway Timeout}, \texttt{emporarily unavailable}.
Fixable: \texttt{must not be empty}, \texttt{cannot be empty}, \texttt{has not been read}, \texttt{No replacement performed}, \texttt{Invalid insert\_line}, \texttt{was omitted}, \texttt{did not match}, \texttt{not unique}, \texttt{verification failed}, \texttt{Invalid arguments}, \texttt{Invalid parameters}.

For the generic rule and MCP results, wider lists follow.
A command's own timeout (\texttt{ommand timed out}, \texttt{ommand timeout}, \texttt{xecution timed out}) stays ambiguous.
Unrecoverable: \texttt{Unauthorized}, \texttt{PERMISSION\_DENIED}, \texttt{UNAUTHENTICATED}, \texttt{ccess denied}, \texttt{ot authorized to}, \texttt{ncorrect API key}, \texttt{nvalid API key} (HTTP~401, gRPC status codes, cloud and API-key errors).
Transient: \texttt{timed out}, \texttt{TimeoutError}, \texttt{ReadTimeout}, \texttt{ConnectTimeout}, \texttt{WriteTimeout}, \texttt{PoolTimeout}, \texttt{socket.timeout}, \texttt{Request Timeout}, \texttt{DEADLINE\_EXCEEDED}, \texttt{UNAVAILABLE}, \texttt{RESOURCE\_EXHAUSTED}, \texttt{onnection reset}, \texttt{onnection refused}, \texttt{onnection aborted}, \texttt{Bad Gateway}, \texttt{Internal Server Error}, \texttt{socket hang up}, \texttt{fetch failed}, \texttt{Max retries exceeded} (Python, \texttt{requests}, \texttt{httpx} and \texttt{urllib3} exceptions, retryable gRPC codes, HTTP 408, 500 and 502, which the OpenAI and Anthropic SDKs retry by default, and Node.js network errors).
Fixable: \texttt{unexpected keyword argument}, \texttt{required positional argument}, \texttt{positional arguments but}, \texttt{nvalid params}, \texttt{INVALID\_ARGUMENT}, \texttt{Bad Request}, \texttt{Unprocessable}, \texttt{validation error}, \texttt{is a required property}, \texttt{is not of type}, \texttt{dditional properties are not allowed}, \texttt{is not a valid tool}, \texttt{nknown tool}, \texttt{ethod not found}, \texttt{-32601}, \texttt{-32602} (Python argument errors, JSON-RPC 2.0, gRPC, HTTP 400 and 422, pydantic and jsonschema messages, LangGraph and the MCP Python SDK).
Texts used for both temporary and permanent conditions, such as \texttt{not available}, and HTTP~404 are left out.

\section{\rev{Implementation Details}}
\label{app:implementation}
\subsection{\rev{Detection}}
\label{app:detect}
Detection favors precision over recall.
Guidance on a result that is not a failure, or the wrong guidance on a real failure, can hurt: in Study~1, moving the failure out of the tool message without guidance made two 7--8B models stop calling on unrecoverable errors far less often (Section~\revblue{\ref{sec:results1}}).
The layer therefore attaches guidance only where a rule fires and the kind is not ambiguous.
It uses three sources of evidence, in order.

\paragraph{Explicit flags.}
A tool message is a failure if the client marks it: \texttt{is\_error} on an OpenAI-style tool message or an Anthropic tool result, \texttt{"isError": true} in an MCP result, \texttt{"success": false} in a Cline result, or an OpenCode tool-execution error object.
These flags are read for every client.

\paragraph{Agent rules.}
For four coding agents identified by their User-Agent (Qwen Code, Codex, Kilo Code and OpenCode), the layer applies text rules recorded from real sessions: for example, Qwen Code's \texttt{File not found:} and a shell result whose \texttt{Exit Code} is non-zero, or Codex's \texttt{Process exited with code N}.

\paragraph{Generic rules.}
For any other client, a tool result is a failure if, after leading white space, it starts with a standard error text produced by a common language, library or protocol.
The rules were written from those sources, not from the evaluation benchmarks, and were frozen before Study~3 was run: Python tracebacks and exception names (\texttt{TypeError:}, \texttt{TimeoutError(}), the \texttt{Error:} prefix of LangChain's and LangGraph's tool nodes, the MCP Python and TypeScript SDK error prefixes, the OpenAI Agents SDK default failure text, HTTP status texts from \texttt{urllib}, \texttt{requests} and \texttt{httpx}, and a JSON object whose \texttt{error} field is a non-empty string or an object.
A JSON \texttt{error} field that is \texttt{null}, \texttt{false} or empty, and error words in the middle of a result, are not failures.

\subsection{\rev{Classification}}
\label{app:classify}
The kind is decided from words in the first 4\,KB of the error text, in priority order: unrecoverable before transient before fixable; anything else is ambiguous.
Unrecoverable words include \texttt{EACCES}, \texttt{EPERM}, \texttt{permission denied}, \texttt{read-only}, \texttt{Unable to write} and \texttt{Forbidden}.
Transient words include \texttt{ECONNRESET}, \texttt{ECONNREFUSED}, \texttt{ETIMEDOUT}, \texttt{Too Many Requests}, \texttt{rate limit}, \texttt{Service Unavailable} and \texttt{Gateway Timeout}.
Fixable words are usage errors that name the value to change, such as \texttt{must not be empty}, \texttt{has not been read}, \texttt{No replacement performed} and \texttt{Invalid parameters}.
For generic clients and MCP results, wider lists add timeouts (\texttt{timed out}, \texttt{DEADLINE\_EXCEEDED}), argument errors (\texttt{unexpected keyword argument}, \texttt{validation error}, JSON-RPC \texttt{-32602}), unknown tools (\texttt{Unknown tool}, \texttt{-32601}) and authorization errors (\texttt{Unauthorized}, \texttt{PERMISSION\_DENIED}).
A shell command's own timeout stays ambiguous, because repeating the same command may time out again; texts that are used both for temporary and permanent conditions, such as \texttt{not available} and HTTP 404, are left out.
The complete lists and their sources are in Appendix~\ref{app:rules}.

\subsection{\rev{Engine and deterministic execution}}
The layer is part of GeulPU \citep{woo2026geulpu}, an open-source inference engine for GGUF models with CUDA and Vulkan back ends\revb{, referred to as the engine below}.
Its engine, GPU kernels and server are written in geul, a programming language with Korean keywords, and built by that language's own compiler; the source code is at \url{https://github.com/wwoosshh/GeulPUv2}, and releases are archived on Zenodo.
Its arithmetic is fixed by a written precision contract with a CPU reference implementation, so a request yields the same tokens and logits whatever the batching, the prompt cache or speculative decoding; in the engine's exact mode, which Studies~1 and~2 used, outputs are also identical across NVIDIA (CUDA) and other (Vulkan) GPUs.
In particular, the same request produces the same tokens on every run, which every study \revblue{verifies} by rerunning a subset of conditions.

\section{\rev{Complete Experimental Protocols and Deviations}}
\label{app:protocols}
\subsection{\rev{Study 1: Where and How to Deliver a Tool Error}}
\label{app:study1}

Study~1 asks where a tool error should be placed, and whether one guidance sentence helps.

The study crosses placement with guidance.
Placement A1 leaves the error in the tool message, as clients send it; placement A2 removes the failed call and its result and restates the error as a runtime note at the end of the user turn.
Guidance B1 adds nothing; B2 adds the sentence of the error's kind (Table~\ref{tab:kinds}).
In the terms of Section~\ref{sec:dimensions}, A1B1 is C0, A1B2 is C1 and A2B2 is C2; A2B1 is the note without guidance.
The design, the stimuli, the analysis script and the hashes of 37 files were registered on 26 October 2026 before any data were collected, and the analysis ran unchanged; there were no deviations.

\paragraph{Stimuli.}
There are 60 synthetic scenarios over six read-only tools (flight status, orders, parcels, books, rooms and menus; ten scenarios each), with fictional values that overlap neither an earlier study of ours nor the pilot.
Every conversation gives the six tool definitions in the same order and no system message.
Each scenario is run with three kinds of error (transient, fixable and unrecoverable) in each of the four forms, which gives 720 conversations per model.
The error texts are new: five variants per kind and tool.

\paragraph{Models and engine.}
Six open models in GGUF format: Qwen2.5-7B-Instruct and Qwen3-8B (Q8\_0), and Llama-3.1-8B-Instruct, Gemma~4 12B, Qwen3.5-9B and the mixture-of-experts Qwen3.6-35B-A3B (Q4\_K\_M).
All run greedily in \revb{the engine's} exact mode with a 4,096-token context.

\paragraph{Measures.}
The primary measure is the correct action at the decision step (Table~\ref{tab:kinds}), with tool calls read leniently: small format slips that a client would still accept are accepted.
Each conversation then continues as an episode of up to five turns with simulated tools.
A transient or fixable episode succeeds when it ends in a text answer that contains the key from the tool's eventual result; an unrecoverable episode succeeds when the first turn is a text answer.
Futile calls are repeated calls to the failed tool after an unrecoverable error.

\paragraph{Hypotheses.}
For transient and fixable errors, \revb{H1.1} predicts A2B1 $>$ A1B1 (placement alone) and \revb{H1.2} predicts A1B2 $>$ A1B1 (guidance alone).
Each is tested per model with an exact McNemar test on 120 pairs (60 scenarios $\times$ two kinds) and is supported if the difference is positive with $p < 0.05$ in at least four of the six models.
\revb{H1.3} asks, two-sided, whether placement still matters once guidance is given (A2B2 vs.\ A1B2), and is decided if at least four models differ in the same direction.
\revb{H1.4} requires that no form lowers correct stopping on unrecoverable errors by more than 10 percentage points against A1B1: the lower bound of a 95\% bootstrap interval (scenarios resampled 10,000 times, pooled over models) must lie above $-10$ points.
A registered rule then selects one form: A1B2 if it passes \revb{H1.2} and \revb{H1.4} and comes within 5 points of A2B2 on transient and fixable errors; otherwise the better of A2B1 and A2B2 among those that pass \revb{H1.4}.
\revb{H1.5} compares the selected form with A1B1 over whole episodes: task completion must be higher in at least four models, and the pooled difference in futile calls must have an upper bound below $+0.5$ calls.

\subsection{\rev{Study 2: An Append-Only Guidance Turn}}
\label{app:study2}

Study~1 selected a form that removes the failed exchange, yet its follow-up showed that one model decides from that exchange, and the real sessions hinted at a cost of removing it.
Study~2 tests the alternative that keeps the history and adds the note in a new user turn (C3) against the selected form (C2), on fresh stimuli and on real agent histories.

Five conditions share the same note and guidance text, byte for byte (Table~\ref{tab:study2forms}).
C3 against C4 isolates the removal of the failed exchange; C2 against C4 isolates the position of the note on synthetic items, where C2 appends it to the user request. C3 against C1 is not a pure position contrast: besides moving the guidance from the tool message to a user turn, C3 adds the note, which restates the call, its arguments and the error.

\paragraph{Registrations.}
The hypotheses came from Study~1's post-hoc follow-up, so they were tested on new stimuli.
The plan was written on 27 October 2026; its second version, with the analysis script and the hashes of 61 files, was registered on OSF\footnote{\url{https://osf.io/7gujd}} before any data existed.
Five deviations were made before measuring and are listed in the registration: Mistral-7B-Instruct-v0.3 was dropped, because its GGUF template has no tool support and its tool template rejects the added user turn of C3 and C4; three Qwen Code sessions were recorded to obtain enough real points; C4 was not run on real histories, where it produces the same messages as C2; the date printed by Llama templates was fixed; and the template server stopped copying weights into RAM.
A third registration extended the study to larger models with the same scripts; it was fixed by the hashes of 45 files before its measurement started, but, by the first author's decision, it was not deposited with a public registry, so we report it as a plan without an external timestamp.

\paragraph{Stimuli.}
The synthetic part has 60 new scenarios over six new tools, each with the three kinds of error in the five conditions: 900 conversations per model.
The real part uses 30 points from six Qwen Code sessions (20 file reads, 7 test runs and 3 directory listings), chosen by a rule fixed before three of the sessions were recorded.
At each point we injected the three kinds of error in Qwen Code's own result format\revb{,} for a file read, \texttt{EBUSY}, a file that moved to a stated new path, and a policy denial (\texttt{EPERM}); for a shell command, \texttt{spawn EAGAIN}, a missing \texttt{python} with the command to use instead, and a sandbox that blocks execution.
This gives 360 real items (30 points $\times$ 3 kinds $\times$ C0--C3).
Because the agent often issued several calls at once, a response is counted as correct if \emph{any} of its calls is the correct action, and the 30 points fall into 19 assistant messages (clusters).

\paragraph{Models.}
The small tier has seven models: Qwen2.5-7B-Instruct, Qwen3-8B, Llama-3.1-8B-Instruct, Gemma~4 12B, Qwen3.5-9B, Qwen3.6-35B-A3B and Llama-3.2-3B-Instruct.
The real items run on Qwen3.6-35B-A3B, the model that produced the sessions.
The large tier has six models: Qwen2.5-32B-Instruct and Qwen3-32B (Q8\_0), and Llama-3.3-70B-Instruct, Gemma~4 31B, Qwen3.5-27B and Qwen3.6-27B (Q4\_K\_M); its real items run on Qwen3.6-27B.
All runs are greedy in \revb{the engine's} exact mode, the small tier on our PC and the large tier on the H100 server.

\paragraph{Hypotheses.}
H2.1: C3 is not worse than C2 on synthetic errors, pooled over models and kinds, with a margin of 5 points (one-sided bootstrap over scenarios).
H2.2: on real histories, C3 gives more correct actions than C2 (exact McNemar, two-sided, with C3 higher).
H2.1 and H2.2 are decided together by Holm's procedure at $\alpha = 0.05$. The registrations name them P1 and P2 (P1-L and P2-L for the large tier).
Secondary hypotheses: S1, removal neither helps nor hurts (90\% interval of C4 $-$ C3 within $\pm 5$ points); S2, guidance in a new turn beats guidance in the tool message (C3 $>$ C1 on transient and fixable pairs in at least five of the seven models); S3, C2 against C4 (descriptive); S4, for Llama-3.1, C3 stops on unrecoverable errors where C2 does not (lower bound of S(C3) $-$ S(C2) above 0); S5, over episodes, task completion is not worse ($-5$ points) and futile calls are not higher ($+0.5$ calls); S6, the prompt tokens recomputed at the decision step (descriptive).

\subsection{\rev{Study 3: Real-Use Evaluation Across Engines}}
\label{app:study3}

Studies~1 and~2 measured the decision step in one engine.
Study~3 asks whether the layer, as shipped, helps in tool-using work: against three widely used engines that have no such layer, against two client-side alternatives, on two public benchmarks (layer~1) and with a real coding agent (layer~2).

The plan, the analysis script (tested on 13 fake-data cases) and the run scripts were fixed by hashes in a local log on 28 October 2026, after a pilot on each machine and before the main measurement; they were not deposited with a public registry.
The detection rules of Section~\ref{sec:detect} had been frozen earlier, before ToolMaze was examined.

\paragraph{Engines and conditions.}
G-off is \revb{the engine} at the commit that froze the detection rules (36e3957; 88555c2 with the parser fix of deviation~1 for Qwen2.5-7B), with the layer off, and G-on the same engine with C3. Both commits precede release 0.15.0, the version cited for the engine, which adds the fix of Appendix~\ref{app:robustness}; the study records \citep{woo2026study} hold the source of both commits and list the builds of Studies~1 and~2 by SHA-256.
L is llama.cpp b11951 with its Jinja templates, V is vLLM~0.32.0 \citep{kwon2023efficient} with automatic tool choice, and S is SGLang~0.5.22 \citep{zheng2024sglang} with its tool-call parser.
Two client-side alternatives run on G-off; they differ from G-on not only in where recovery happens but also in what they know and do.
E1 appends the guidance sentence to the result of the injected fault, with the fault's exact kind: the best case for a client hook such as an ADK callback, a LangChain middleware or an MCP interceptor, which knows what it injected.
E2 retries a transient fault up to three times inside the tool layer, as a LangChain-style tool retry does.
All engines sit behind the same proxy, which sets temperature 0 and the same chat-template options; per model, all engines use the same context and concurrency.

\paragraph{Models.}
The small tier has Llama-3.1-8B-Instruct, Qwen3-8B and Qwen2.5-7B-Instruct, the models for which Study~2 found a large ($+27.8$ points), a small ($+1.1$) and a negative ($-2.8$) effect of C3.
\revb{The engine} and llama.cpp run the Q8\_0 files on our PC; vLLM and SGLang run the BF16 weights on the H100.
The large tier, on the H100, has Qwen3-32B (Q8\_0; FP8 for vLLM and SGLang) and Llama-3.3-70B-Instruct (Q4\_K\_M; \revb{the engine} and llama.cpp only), \rev{respectively, a model with little or no delivery effect and the model with the largest effect in Study~2\textquotesingle s large tier.}
Quantization differs between engines; \revb{H3.2} below is meant to reveal its effect.

\paragraph{Layer 1: benchmarks.}
Loud/Quiet \citep{kraishan2026loud,kraishan2026brittle} runs 24 multi-turn BFCL tasks under five conditions\revb{,} clean, a timeout, silent corruption, schema drift and a missing tool\revb{,} and scores the agreement of the final environment state with the reference (0--1); its loop, system text and step limit are unchanged.
We had seen its error texts before writing the rules, which classify its timeout as transient, schema drift as fixable and the missing tool as ambiguous (no guidance).
ToolMaze \citep{zhu2026when,zhu2026toolmaze} runs 100 tasks with one failure point in three modes\revb{,} no fault (P0), a transient fault (P1) and a permanent fault (P2)\revb{,} and passes or fails each run with its own rule-based judge; the large tier ran the first 50 tasks per mode. We used its code at commit f8620d8 and data revision 08b0239, with the first of its four complexity levels. Its judge passes a P0 run if every required tool succeeds, a P1 run if the model calls the failed tool at least twice, one call succeeds and no required tool is missing, and a P2 run if the model reaches the failing tool and then calls no other tool; a P2 pass therefore rewards stopping, not completion of the original task.
ToolMaze was first examined after the rules were frozen: of its 1,176 explicit errors, the rules attach the correct guidance to 29.8\%, the wrong guidance to 3.9\% (a permanent error read as transient) and none to 66.3\%. Among the errors that received guidance, 11.6\% received the wrong kind.
A run that an engine ends with an error, such as llama.cpp's HTTP~500 when it cannot parse a tool call, scores 0, because that is what a user experiences; such failures are also counted. Absolute ToolMaze pass rates with G-off and G-on were 61.0\% and 60.0\% without a fault, 2.7\% and 28.3\% under a transient fault and 48.3\% and 51.0\% under a permanent fault in the small tier (three models, 100 tasks per mode), and 83.0\% and 86.0\%, 45.0\% and 55.0\%, and 51.0\% and 62.0\% in the large tier (two models, 50 tasks per mode). Loud/Quiet's state agreement on explicit errors was 80.2\% and 79.1\% in the small tier and 77.4\% and 77.9\% in the large tier.

\paragraph{Hypotheses.}
Each tier and benchmark is analyzed separately, with a bootstrap over tasks (10,000 resamples).
\revb{H3.1}: on explicit errors (Loud/Quiet's timeout, schema drift and missing tool; ToolMaze's P1 and P2), G-on beats each of L, V and S (one-sided, Holm over the three comparisons).
\revb{H3.2}: G-off starts from the same baseline as each engine: the 90\% interval of G-off $-$ X over all conditions lies within $\pm 5$ points; otherwise \revb{H3.1}'s gain is reported as mixed with engine differences.
\revb{H3.3}: G-on against E1 and E2 (descriptive). \revb{H3.4}: on Loud/Quiet's missing tool, G-on stops calling the hidden tool more often than each engine (descriptive).
A control compares G-on with G-off where the layer should not act: Loud/Quiet's clean and silent-corruption conditions and ToolMaze's P0.

\paragraph{Layer 2: a real coding agent.}
Qwen Code~0.25.3 works on six small repositories.
Each task fetches data from a data service with \texttt{python fetch\_data.py}, writes a result file, and checks it with \texttt{pytest}; the expected values are computed from the data, so a fabricated result fails.
The data service is healthy (ok), fails the first two requests with HTTP~503 (flaky), or always answers 403 with a message that a token is required (forbidden).
The agent sees the error texts that Python's \texttt{urllib} prints, and the frozen rules classify them, in Qwen Code's shell-result format, as transient and unrecoverable.
A run succeeds if the tests pass (ok and flaky) or, for forbidden, if the agent's final message reports the access problem; we also count requests to the data service and whether a result file was written, which under forbidden can only be fabricated. Because the report is detected by keywords, we read all 24 final messages of the forbidden runs on the engine: each of the 12 with the layer states the 403 access error and claims no success, although one also guesses that the server may not be running; of the seven without the layer that the keywords counted as reports, one claims that the tests passed after an authorization header was added, one only lists possible causes, and two report the error but leave a fabricated result file.
Qwen3-8B (concurrency 4) and Qwen3-32B (concurrency 1) run with a 32,768-token context on all five engines, once per task and fault; with six tasks, layer~2 is descriptive.

\paragraph{Deviations.}
We report six deviations, all recorded with hashes in the log.
(1)~Qwen2.5-7B's GGUF template shows the tool-call example in doubled braces; the model copied them and \revb{the engine} did not parse its calls (none in 1,734 replies), while llama.cpp's parser recovered them.
We fixed the parser and re-ran Qwen2.5-7B's \revb{rows on the engine}; the old rows are kept.
(2)~While checking the field names of ToolMaze's judge output, we saw one pass value of one pilot trial; pilot runs are not analyzed.
(3)~The registration is local, as stated above.
(4)~In layer~2, SGLang rejected every request (HTTP~400): Qwen Code asks for 32,000 output tokens, which with the prompt exceeds the context, and the other engines clamp the request.
We added a run of SGLang with the limit set to 8,192 tokens by the proxy.
It is not the same condition: with Qwen3-32B, four answers of other engines in the main run were longer than 8,192 tokens (all ran to the context limit), and SGLang still rejects prompts longer than 24,576 tokens, which the others accepted.
We report both SGLang runs.
(5)~Layer~2's working folders were named after the engine, so only \revb{the engine's} contained Hangul characters, which the models had to copy into absolute file paths.
Qwen3-32B's tool calls broke off at the first such character (12 of 18 G-on runs and 3 of 18 G-off runs), and Qwen3-8B misspelled the path (10 of 18 G-on runs).
We re-ran the \revb{engine} rows with ASCII names, with the same build and settings, and report the confounded rows separately \rev{(Appendix~\ref{app:robustness} discusses the engine-side cause).}
(6)~ToolMaze's judge passes a P1 run only if the model itself calls the failed tool at least twice, so a retry inside the tool layer (E2) fails by construction; we interpret E2 on Loud/Quiet only.

\section{\rev{Complete Study 1 Decision-Step Results}}
\begin{table}[H]
\caption{Study~1: share of correct actions at the decision step (60 conversations per cell; greedy). A1: error in the tool message; A2: runtime note in the user turn with the failed call removed; B1: no guidance; B2: one guidance sentence. The last two rows are the larger-model extension on an H100.}
\label{tab:study1}
\footnotesize
\setlength{\tabcolsep}{3pt}
\begin{tabular}{@{}lcccccccccccc@{}}
\toprule
& \multicolumn{4}{c}{Transient (repeat)} & \multicolumn{4}{c}{Fixable (fix and repeat)} & \multicolumn{4}{c}{Unrecoverable (stop)} \\
\cmidrule(lr){2-5}\cmidrule(lr){6-9}\cmidrule(l){10-13}
Model & A1B1 & A1B2 & A2B1 & A2B2 & A1B1 & A1B2 & A2B1 & A2B2 & A1B1 & A1B2 & A2B1 & A2B2 \\
\midrule
Qwen2.5-7B & .37 & .43 & 1.00 & 1.00 & .48 & .87 & 1.00 & 1.00 & 1.00 & 1.00 & .30 & 1.00 \\
Qwen3-8B & .00 & .08 & 1.00 & 1.00 & .02 & 1.00 & .95 & 1.00 & 1.00 & 1.00 & .60 & 1.00 \\
Llama-3.1-8B & .98 & .98 & 1.00 & 1.00 & .98 & .93 & .92 & 1.00 & .02 & .10 & .00 & .00 \\
Gemma~4 12B & .12 & .23 & .95 & 1.00 & .80 & 1.00 & 1.00 & 1.00 & 1.00 & 1.00 & 1.00 & 1.00 \\
Qwen3.5-9B & .30 & .97 & 1.00 & 1.00 & .13 & .87 & 1.00 & 1.00 & 1.00 & 1.00 & .93 & 1.00 \\
Qwen3.6-35B-A3B & .33 & 1.00 & 1.00 & 1.00 & .38 & 1.00 & .87 & .98 & 1.00 & 1.00 & 1.00 & 1.00 \\
\midrule
Qwen2.5-32B & .42 & .78 & .97 & 1.00 & .33 & .98 & .93 & 1.00 & 1.00 & 1.00 & .98 & 1.00 \\
Qwen3-32B & .32 & .55 & .92 & 1.00 & .23 & 1.00 & .98 & 1.00 & 1.00 & 1.00 & 1.00 & 1.00 \\
\bottomrule
\end{tabular}
\end{table}

\section{\rev{Complete Study 3 Benchmark Comparisons}}
\begin{table}[H]
\caption{Study~3, layer~1: differences in percentage points (Loud/Quiet: state agreement; ToolMaze: pass rate). The first row is descriptive, with 95\% intervals. \revb{H3.1} rows: G-on $-$ X on explicit errors (* supported after Holm). \revb{H3.2} rows: G-off $-$ X over all conditions (= baselines equal within $\pm 5$ points at 90\%, $\neq$ not shown equal). \revb{Small tier: Llama-3.1-8B, Qwen3-8B and Qwen2.5-7B; large tier: Qwen3-32B (the only model for V and S) and Llama-3.3-70B.}}
\label{tab:study3l1}
\scriptsize
\setlength{\tabcolsep}{2.5pt}
\begin{tabular}{@{}lllll@{}}
\toprule
& \multicolumn{2}{c}{\revb{Small tier}} & \multicolumn{2}{c}{\revb{Large tier}} \\
\cmidrule(lr){2-3}\cmidrule(l){4-5}
& Loud/Quiet & ToolMaze & Loud/Quiet & ToolMaze \\
\midrule
G-on $-$ G-off, explicit errors & $-1.1$ $[-2.8, 0.3]$ & $+14.2$ $[10.0, 18.3]$ & $+0.5$ $[-0.4, 1.6]$ & $+10.5$ $[4.0, 17.0]$ \\
\revb{H3.1}: G-on $-$ L & $+15.3$* & $+15.7$* & $+0.7$ & $+30.0$* \\
\revb{H3.1}: G-on $-$ V & $+14.4$* & $+13.9$* & $+1.2$ & $+9.0$ ($p = 0.052$) \\
\revb{H3.1}: G-on $-$ S & $+12.8$* & $+14.7$* & $+2.9$* & $+12.0$* \\
\revb{H3.2}: G-off $-$ L & $+17.0$ $\neq$ & $+4.2$ $\neq$ & $+0.9$ = & $+23.3$ $\neq$ \\
\revb{H3.2}: G-off $-$ V & $+16.0$ $\neq$ & $-0.5$ = & $+0.5$ = & $+0.7$ $\neq$ \\
\revb{H3.2}: G-off $-$ S & $+14.0$ $\neq$ & $+0.6$ = & $+0.6$ = & $+2.7$ $\neq$ \\
\revb{H3.3}: G-on $-$ E1 & $-1.1$ $[-2.6, 0.2]$ & $-16.8$ $[-22.0, -11.8]$ & $+1.4$ $[-0.2, 3.7]$ & $-18.5$ $[-26.0, -11.0]$ \\
\revb{H3.3}: G-on $-$ E2 (transient) & $-3.5$ $[-8.7, 0.6]$ & (not valid) & $-1.0$ $[-4.1, 2.4]$ & (not valid) \\
Control: G-on $-$ G-off & $-0.3$ $[-0.8, 0.0]$ & $-1.0$ $[-4.7, 2.7]$ & $0.0$ $[0.0, 0.0]$ & $+3.0$ $[-1.0, 7.0]$ \\
\bottomrule
\end{tabular}
\end{table}

\section{\rev{Complete Coding-Agent Comparisons}}
\begin{table}[H]
\caption{Study~3, layer~2 (Qwen Code, Qwen3-8B and Qwen3-32B pooled; 12 pairs over six tasks; differences in percentage points with 95\% bootstrap intervals). Flaky: tests pass. Forbidden: the agent reports the access problem; requests: data-service requests per run. S: SGLang's main run (all requests rejected) and the added run with 8,192 output tokens.}
\label{tab:study3l2}
\scriptsize
\setlength{\tabcolsep}{2pt}
\begin{tabular}{@{}llllll@{}}
\toprule
& vs.\ G-off & vs.\ L & vs.\ V & vs.\ S (main) & vs.\ S (8,192) \\
\midrule
Flaky: G-on $-$ X & $+17$ $[0, 33]$ & $+42$ $[25, 50]$ & $0$ $[-25, 25]$ & $+67$ $[50, 83]$ & $+33$ $[-8, 67]$ \\
Forbidden, reports: G-on $-$ X & $+42$ $[8, 75]$ & $+25$ $[8, 42]$ & $+33$ $[8, 67]$ & $+100$ & $+42$ $[17, 67]$ \\
Forbidden, requests: G-on $-$ X & & $-4.0$ $[-6.0, -1.9]$ & $-4.5$ $[-6.4, -2.3]$ & $+1.0$ & $-2.0$ $[-3.1, -0.8]$ \\
Ok, baseline: G-off $-$ X & & $-8$ $[-58, 42]$ & $+8$ $[-25, 42]$ & $+58$ $[25, 92]$ & $0$ $[-33, 33]$ \\
Ok, control: G-on $-$ G-off & $-8$ $[-50, 25]$ & & & & \\
\bottomrule
\end{tabular}
\end{table}

\section{\rev{Detailed Study 1 Results and Follow-Ups}}
\label{app:results1}
\paragraph{Registered verdicts.}
\revb{H1.1} was supported in five of six models; the discordant pairs ran, for example, 116 to 0 for Qwen3-8B ($p = 2.4\times10^{-35}$), while Llama-3.1-8B, already at 0.98, had 2 to 5 ($p = 0.45$).
\revb{H1.2} was supported in five of six models as well, but its effect depended on the model: a guidance sentence in the tool message lifted Qwen3.5 and Qwen3.6 to near 100\% on transient errors, while Qwen2.5, Qwen3 and Gemma~4 hardly moved (0.37 to 0.43, 0.00 to 0.08, 0.12 to 0.23).
Placement still mattered after guidance (\revb{H1.3}): A2B2 was better than A1B2 in four models (all significant), and no model differed significantly the other way.
On unrecoverable errors, A1B2 ($+1.4$ points, 95\% interval $[0.0, 2.8]$) and A2B2 ($-0.3$, $[-0.8, 0.0]$) passed \revb{H1.4}, but A2B1 did not ($-19.7$, $[-23.3, -16.1]$): without guidance, the note made Qwen2.5-7B and Qwen3-8B call again where they should have stopped (stopping fell from 1.00 to 0.30 and 0.60).
The selection rule rejected A1B2, whose pooled correct action on transient and fixable errors (78.1\%) was more than 5 points below A2B2's (99.9\%), and chose A2B2, the only note form that passed \revb{H1.4}.
The engine's form C2 is this selected form.

\revb{H1.5} was not supported.
Task completion with A2B2 was higher than with A1B1 in five models\revb{,} for example, Qwen3-8B completed 0\% of transient episodes with A1B1 and 92\% with A2B2\revb{,} but the pooled difference in futile calls was $+0.56$ calls ($[0.51, 0.61]$), above the margin.
All of it came from Llama-3.1-8B ($+3.35$ calls; $0.00$ for the other five): with A2B2 it called the same tool until the turn limit in 55 of 60 unrecoverable episodes.
A 10\% rerun (72 conversations per model) reproduced the tokens of every turn exactly, and parsing failures at the decision step were 1 of 720 for Llama-3.1-8B and 0 for the other models.

\paragraph{Larger models.}
A second registration extended the study to Qwen2.5-32B-Instruct and Qwen3-32B (Q8\_0) on the H100 server, with the analysis script fixed before the results were opened.
Before measuring, three kernel modules were byte-identical to those built on our development PC, and 24 conversations reproduced, token for token, the output of an RTX~4070~Ti on Windows.
The conclusions held at 32B: A2B2 beat A1B1 on transient and fixable errors (75 to 0 and 87 to 0 discordant pairs), reached 1.00 in every transient and fixable cell, and kept stopping at 1.00.
The client's form improved with size\revb{, with Qwen3 rising} from 0.00 at 8B to 0.32 at 32B on transient errors\revb{,} but still missed most corrections, and the guidance sentence in the tool message (A1B2) reached only 0.55--0.78 on transient errors.

\paragraph{Why Llama 3.1 does not stop.}
Two registered follow-ups examined Llama-3.1-8B's behavior on unrecoverable errors.
Removing the tool instruction that its chat template places in the first user message did not change it (stopping under A2B2 stayed at 0 of 60).
The second follow-up varied what follows the failure and measured, at the decision step, the log-odds of starting a call rather than text.
No added content made Llama stop: the runtime note, the guidance sentence, the same text written by the user, the note in a system message, and even ``\textit{Do not call any tool again.}'' all left stopping at 0 of 60, and the variants in the user turn moved the decision slightly \emph{towards} a call.
What stopped it was the structure of the history: two or three of its own failed exchanges (41 and 58 of 60), or, after one failed exchange, a new user message (59 of 60).
The A2 note therefore removes the signal that Llama-3.1-8B uses\revb{,} its own failed call in the history\revb{,} and keeps only the one it ignores.
A form that keeps the failed exchange and opens a new user turn after it is C3, which Study~2 tests.

\paragraph{Real sessions.}
As an exploratory check, the four forms were applied at three points of real Qwen Code sessions with fixable errors (a failed \texttt{cd} in Windows shells) for Qwen3.6-35B-A3B.
The client's form was already correct at all three; A2B2 missed two, replacing the command with a different one.
Three points do not support a verdict, but they show that a removed history can cost context in a long session\revb{, which motivated testing an append-only form}.

%% ===================================================================

\section{\rev{Detailed Study 2 Results}}
\label{app:results2}
\paragraph{Small tier.}
H2.1 was supported: pooled over 1,260 pairs, C3 was 3.9 points better than C2 ($p = 10^{-4}$).
Most of the difference came from Llama-3.1-8B ($+27.8$ points; the other six models ranged from $-2.8$ to $+1.1$), the model that C2 did not stop: on unrecoverable errors it stopped in 98\% of cases with C3 and in none with C2 or C4 (S4).
C3 was not better everywhere: for Qwen2.5-7B it was 2.8 points worse ($[-5.6, -0.6]$), and on transient errors Llama-3.1-8B repeated the call in 82\% of cases with C3 against 100\% with C2.
H2.2 was not supported: on the 90 real items C3 was correct in 55.6\% and C2 in 51.1\% (17 against 13 discordant pairs, $p = 0.58$), although C3 beat the client's form by 34.4 points.
\rev{Removal met the registered equivalence criterion (S1), rather than showing an exactly zero effect}, C2 and C4 agreed on all but one pair (S3), and the note with guidance in a new turn (C3) beat guidance in the tool message (C1) in five of seven models (S2)\revb{,} with Llama-3.1-8B again the exception (11 against 2 pairs in favor of C1).
Over whole episodes, C3 completed slightly more tasks than C2 and made 0.71 fewer futile calls (S5).

\paragraph{Large tier.}
Both primary hypotheses were supported, also after Holm's correction.
H2.1: C3 was 5.3 points better than C2 over 1,080 pairs, again driven by Llama ($+32.8$ points for Llama-3.3-70B; between $-1.1$ and $0$ for the other five).
The defect of C2 persisted at 70B: on unrecoverable errors Llama-3.3-70B stopped in 95\% of cases with the client's form, in 2\% with C2 and in 100\% with C3.
H2.2: on real histories, Qwen3.6-27B was correct in 57.8\% of items with C3 and in 45.6\% with C2 (16 against 5 discordant pairs, $p = 0.027$); a bootstrap over the 19 clusters gave $+12.2$ points ($[4.0, 20.2]$).
A descriptive comparison of the tiers found that only the Llama family's advantage for C3 grew with size ($+5.0$ points, $[1.1, 8.9]$); the real-history difference grew by $+7.8$ points with a wide interval ($[-3.0, 20.8]$). \paragraph{Sensitivity of H2.2 (exploratory).} The 90 real-history pairs of each tier come from 30 failure points in 19 assistant messages of six sessions, so they are not independent. In the large tier every session favored C3 (by 5.6 to 33.3 points); sign-flip permutation tests over the 19 messages and over the six sessions gave $p = 0.028$ and $p = 0.031$, and leaving out one session at a time gave McNemar $p$ between 0.031 and 0.077. The registered score counts a response as correct if any of its calls is the correct action. With C3, 12 of the 50 responses scored correct by a call also contained other calls, against 7 of 37 with C2, mostly file listings, test runs and reads of other files; counting such responses as incorrect gives 44.4\% against 37.8\% (11 against 5 discordant pairs, $p = 0.21$). In the small tier no analysis approached significance ($p \geq 0.40$). \paragraph{Stopping answers.} An unrecoverable episode counts as stopped when the first turn is a text answer. A search of the 3,261 text answers to unrecoverable errors in both tiers for reported values (times, readings, approvals and grades) found two fabricated meter readings, both with the client's form (Gemma~4 12B and Qwen3.5-9B); in the answers we read, the model restated the cause of the failure, apart from a few refusals that do not mention it, such as two of Llama-3.3-70B's 57 text answers with the client's form.

\paragraph{Cost.}
The registered cache measure (S6) did not favor the append-only form at the decision step.
The median number of prompt tokens recomputed there was 171 for C3 against 94 for C2 on synthetic items, and 1,081 against 965 on real histories (78 and 953.5 for the client's form): C3 keeps the failed result, which C2 drops, and adds the note after it.
An advantage of keeping the history for the prefix cache, as argued by \citet{ji2025context}, would therefore have to come from later turns, which this study did not measure.

\paragraph{Corrective gain.}
As an exploratory measure we computed FTB's corrective gain \citep{gumaan2026feedback} on our synthetic items: the log-probability of the failed call without the failed exchange minus that with it.
Averaged per token, it was $+0.45$ in the small tier and $+0.52$ in the large tier, and in about 12\% of items the failure record made the failed call \emph{more} likely, FTB's feedback inversion.
Gemma~4 could not be scored: its template places the tool result inside the same model turn as the call, so our rule for cutting the candidate call at the end-of-turn marker did not apply.

\rev{Study~2 supports C3 as an aggregate choice under the registered comparisons. It is non-inferior on pooled synthetic errors, avoids C2\textquotesingle s stopping failure for Llama models and is better on the measured real histories in the large tier under the registered scoring rule. It is not best for every model; small-tier real-history superiority is not supported.}
The engine's C3 is the form of this study.

%% ===================================================================

\section{\rev{Engine Robustness and Reported Defects}}
\label{app:robustness}
\paragraph{Engine robustness decided more than delivery.}
The largest differences between engines were not about recovery.
On Loud/Quiet, llama.cpp could not parse Llama-3.1-8B's tool calls (HTTP~500 for 34 of 85 requests in the pilot), and vLLM and SGLang rejected its histories with several calls per turn (HTTP~400); together these ended 120--137 runs per engine.
llama.cpp passed 24\% of fault-free ToolMaze runs with Llama-3.3-70B, against \revb{\revblue{G-off's}} 82\%.
SGLang rendered the same first request with 364 more tokens than the other engines for Llama-3.1-8B and 280 more for Qwen3-32B, and in layer~2 it rejected every request of Qwen Code.
\revb{The engine} had failures of its own.
It did not parse Qwen2.5-7B's doubled braces until we fixed its parser (deviation~1).
And the layer-2 analysis revealed a second defect.
At the point where Qwen3-32B was writing a rare token of a file path, its most likely next token was the special token \texttt{<|im\_start|>}, in both \revb{the engine} (28\%, against 7\% for the correct character) and llama.cpp (21\% against 14\%); the tokenizers of the two engines and of the reference implementation agreed.
llama.cpp, which constrains tool calls with a grammar, kept the call well-formed, with the character missing; \revb{the engine} emitted the token, the model ended its turn, and the agent took the broken call as its final answer.
\revb{The engine's 0.15.0 release} now excludes, while a Qwen-style call is open, special tokens other than the call's closing tag and the end-of-generation tokens from the candidates.
On the broken request this yields the complete, correct call, and over six sampled runs at temperature 0.7 it lost no call where the previous build lost three; requests in which no special token is chosen inside a call are unchanged, bit for bit.
All \revb{of the engine's} rows of Study~3 use the build before this fix.
\rev{Cross-engine comparisons must count parsing failures and verify baseline agreement, as \revb{H3.2} did.}

\end{document}